\documentclass{amsart}

\usepackage{amsmath,amssymb,amsthm}
\usepackage{algorithm}
\usepackage{algorithmic}
\usepackage{multirow}
\usepackage{color}
\usepackage{bm}
\usepackage{graphicx}
\usepackage{lscape}

\DeclareMathOperator{\tr}{tr}
\DeclareMathOperator{\T}{T}
\DeclareMathOperator{\st}{s.t.}

\title[MvOPLS]{Multi-view Orthonormalized Partial Least Squares: Regularizations and Deep Extensions}
\author[Li Wang] {Li Wang}
\author[Ren-Cang Li]{Ren-Cang~Li}
\author[Wen-Wei Lin]{Wen-Wei Lin}
\thanks{
	Li Wang is with Department of Mathematics and Department of Computer Science and Engineering, University of Texas at Arlington, Arlington, TX 76019-0408, USA. Email: li.wang@uta.edu. Corresponding Author.\\
\indent Ren-Cang Li is with Department of Mathematics, University of Texas at Arlington, Arlington, TX 76019-0408, USA. Email: rcli@uta.edu.\\
\indent Wen-Wei Lin is with Department of Applied Mathematics, National Chiao Tung University, Hsinchu 300, Taiwan, R.O.C. 
Centre in Taiwan. Email: wwlin@am.nctu.edu.tw.
}

\begin{document}
\maketitle

\begin{abstract}
We establish a family of subspace-based learning method for multi-view learning using the least squares as the fundamental basis. Specifically, we investigate orthonormalized partial least squares (OPLS) and study its important properties for both multivariate regression and classification. Building on the least squares reformulation of OPLS, we propose a unified multi-view learning framework to learn a classifier over a common latent space shared by all views. The regularization technique is further leveraged to unleash the power of the proposed framework by providing three generic types of regularizers on its inherent ingredients including model parameters, decision values and latent projected points. We instantiate a set of regularizers in terms of various priors. The proposed framework with proper choices of  regularizers not only can recast  existing methods, but also inspire new models. To further improve the performance of the proposed framework on complex real problems, we propose to learn nonlinear transformations parameterized by deep networks. Extensive experiments are conducted to compare various methods on nine data sets with different numbers of views in terms of both feature extraction and cross-modal retrieval.
\end{abstract}

\section{Introduction}\label{sec:introduction}
Data sets are increasingly collected from different views of one object in many real world applications \cite{li2018survey}. This results in depicting,
more comprehensively, the object from multiple views than solely relying on a single view.
Each view is composed of its own set of features. As each sub-data set for one view within the multi-view data contains complementary information of the same object,
it is expected that  learning algorithms should make good
use of these views for the best outcome \cite{foster2008multi}. 

Multi-view learning \cite{xu2013survey,li2018survey} is such a learning mechanism seeking to leverage the complementary information of multiple views to boost learning performance. Many multi-view learning algorithms have been proposed in the literature. Among them, subspace-based learning approaches have attracted much attention. They aim to obtain a common latent  subspace shared by all views under the assumption that these views are generated from the common subspace. The subspace-based learning algorithms have demonstrated a great deal of success in many tasks such as cross-modal retrieval \cite{kan2015multi,sharma2012generalized} and feature extraction \cite{sun2010canonical,sun2015multiview}. 

In this paper, we concentrate on the study of a family of the subspace-based multi-view learning algorithms in terms of the least squares formulation from three different perspectives: (i) two/more views, (ii) linear/nonlinear representation, and (iii) unsupervised/supervised learning. 

The most representative model in multi-view learning is canonical correlation analysis (CCA), which was originally proposed to learn two linear projection matrices by maximizing the correlation between two views in a common space \cite{harold1936relations}. It has since been extended for more than two views \cite{kettenring1971canonical,nielsen2002multiset}, nonlinear projections via either kernel representation \cite{lai2000kernel} or deep representation \cite{andrew2013deep}, and supervised learning \cite{sharma2012generalized}. Moreover, the least squares reformulations of CCA have been proposed for supervised multi-label classification \cite{sun2010canonical} and unsupervised learning of more than two views \cite{via2007learning}. They have demonstrated great advantages in yielding effective models and efficient learning algorithms. However, the reformulation in \cite{sun2010canonical} is essentially a single-view classification method since it treats data point as one view and class label as another. In addition to CCA, other forms of least squares have been studied for two views such as coupled spectral regression  \cite{lei2009coupled}  
and partial least squares (PLS) \cite{li2003multimedia,wold1984multivariate}. 

Least squares formulation has been previously studied for single-view supervised learning, but it is seldom explored for subspace-based multi-view learning. As to single-view learning, linear discriminant analysis (LDA) can be formulated as least squares for both binary classification \cite{bishop2006pattern} and multi-class classification \cite{ye2007least}. CCA for supervised classification is proved to be equivalent to LDA for multiclass classification \cite{hastie1995penalized}, so CCA shares the same least squares reformulation as that of LDA. 
Although CCA and LDA have their own least squares formulations for two views, it is not straightforward to derive the least squares  models for multiple views.
For example, LDA has been generalized to learning projection matrices for binary classification of two views \cite{diethe2008multiview}. Multi-view discriminant analysis (MvDA) further extends LDA for multi-class classification of more than two views \cite{kan2015multi}. Various combinations of CCA and LDA are also proposed in \cite{sharma2012generalized,cao2017generalized,sun2015multiview}. Instead of least squares, these methods are originally modeled as a trace ratio problem, but it is the relaxed ratio trace problem that gets finally solved for the convenience of numerical treatment and, as a result, their solutions are not optimal to their original models  \cite{wang2007trace}. 

In order to characterize various existing supervised subspace-based learning methods, we investigate orthonormalized partial least squares (OPLS) \cite{worsley1997characterizing}, which was proposed to perform dimensionality reduction only in the input space, and that makes it different from and also less popular than CCA and PLS. Precisely this property of reduction in the input space becomes its advantage for multivariate analysis in the setting of supervised learning because the prediction primarily relies on reliable extraction of good features in the input space. The equivalence between supervised CCA and OPLS was established in \cite{sun2009equivalence}. Kernel OPLS was proposed in \cite{arenas2008efficient} for learning nonlinear transformations. As another advantage, OPLS admits a least squares formulation that leads to an optimal regression classifier in a latent space \cite{arenas2008efficient,roweis1999linear}. These methods generally assume the data consists of one input and one output. Hence, they do not appear suitable for supervised multi-view learning.

Building on the least squares formulation of OPLS, we propose a unified multi-view learning framework for  subspace-based learning. The framework aims to learn a classifier over the latent space shared by all multiple views. Various regularizations are presented to enrich the proposed framework, which not only can recast many existing methods, but also inspire new models.  The proposed framework provides a natural way to deal with arbitrary number of views with or without class labels by learning either linear or nonlinear projections. The main contributions of this paper are summarized as follows:

1) We revisit OPLS for multivariate regression analysis and study its properties especially for multi-class classification. We show that LDA is a special case of OPLS for multi-class classification, and regularized LDA is equivalent to the shrinkage estimator of the covariance matrix for LDA.

2) A novel multi-view learning framework, namely regularized multi-view OPLS (MvOPLS), is proposed, and three general purposed regularizations are studied with some examples built on the inherent ingredients of the framework, including model parameters, decision values and latent projected points. A  unified optimization algorithm is also presented via the generalized eigenvalue decomposition.

3) We recast several existing methods with the set of provided regularizers under the proposed framework. New models are also motivated by integrating proper regularizers into MvOPLS. To deepen the understanding of existing methods, we compare them and highlight their differences in terms of the choices of regularizations.

4) We explore the proposed regularized MvOPLS to learn nonlinear transformations parameterized by deep networks. All instantiated methods under the framework can take advantage of the proposed nonlinear extension with little additional effort. This will provide a large set of deep supervised subspace-based multi-view learning methods.

5) Extensive experiments are conducted to compare twenty-seven methods instantiated from the proposed framework on nine data sets with various numbers of views. These methods are evaluated and compared on two different tasks: feature extraction and cross-modal retrieval. Results show that subspace-based learning for a common latent space is effective and its nonlinear extension can further boost the performance.

In the rest of this paper, we first briefly review existing methods related to this work in Section \ref{related-work}. In Section \ref{sec:opls}, we study the properties of OPLS for both regression and classification in single-view learning. In Section \ref{reg:rmvopls}, the proposed multi-view learning framework is detailed. Under the framework, we recast seven existing methods and present two new ones in Section \ref{sec:reformulation}, and further explore their deep variants to learn nonlinear transformations parameterized by deep networks in Section \ref{sec:deep}. Extensive experiments are conducted in Section \ref{sec:experiments}. Finally, we draw our conclusion in Section \ref{sec:conclusion}.

\section{Related Work} \label{related-work}
We briefly review existing methods relevant to this work from the three perspectives mentioned in Section \ref{sec:introduction}. Specially, we discuss methods in two broad categories, i.e., unsupervised and supervised, and then their extensions to more than two views and nonlinear transformations. 

In the setting of unsupervised learning, CCA has been the workhorse for learning a common latent space of two views \cite{harold1936relations}. To deal with more than two views, multiset CCA (MCCA) \cite{kettenring1971canonical,nielsen2002multiset} based on pairwise correlations and generalized CCA (GCCA) \cite{horst1961generalized} by aligning all views via a common representation are proposed. Among them, MCCA with a least squares formulation \cite{via2007learning} is widely used  due to its simplicity. Kernel CCA (KCCA) \cite{lai2000kernel} and deep CCA (DCCA) \cite{andrew2013deep} are two representative approaches to explore nonlinear projections to model complex real world data sets  via the kernel trick and deep learning, respectively.
dMCCA \cite{somandepalli2019multimodal} extends MCCA to nonlinear transformations via deep networks, but it  can only deal with the very special case where all views must reside in the same input space. Deep GCCA (DGCCA) \cite{Benton2019} extends GCCA to nonlinear transformations but it does not reduce to CCA for two views. Other linear models closely related to CCA \cite{arenas2009kernel} have also been explored especially for two views, including PLS \cite{li2003multimedia} and OPLS \cite{sun2009equivalence,arenas2008efficient}, but they are less popular for subspace-based learning. In addition,  spectral regression is used to learn the common space between two views in two separate steps \cite{lei2009coupled}. These methods do not explicitly take into account supervised information such as the class labels for multi-class classification. 

Various supervised subspace-based learning approaches have recently been proposed in order to integrate supervised information to improve multi-view learning. LDA \cite{yan2005graph} is the main tool for  subspace learning with supervised information. The combination of LDA and CCA has been successfully used to find a discriminant subspace.
Generalized multi-view analysis (GMA) \cite{sharma2012generalized} obtains a discriminant common space by incorporating intra-view discrimination information and cross-view correlation, and is directly applicable to learn nonlinear transformations via the kernel trick for more than two views. Different from GMA, MLDA \cite{sun2015multiview} replaces the within-class scatter matrix with the covariance matrix.
Multi-view discriminant analysis (MvDA) \cite{kan2015multi} considers both inter-view and intra-view variations leading to a more discriminative common space. Its nonlinear extension through deep networks has been studied in \cite{kan2016multi}, resulting in a discriminant and view-invariant representation shared among all multiple views. Multi-view modular discriminant analysis (MvMDA) \cite{cao2017generalized} is proposed to maximize the distances between different class centers across different views and minimize the within-class scatter of each view. Most of these methods are originally formulated as trace ratio problems, but solved as relaxed ratio trace problems. Since two types of optimization problems are not equivalent \cite{wang2007trace}, their solutions are not optimal to their original models.

In the following, we will propose a unified multi-view learning framework, which can recast most of the above-mentioned subspace-based learning approaches. It provides a natural and accurate interpretation to the relaxed problems of existing models under a unified framework. The proposed framework combined with  powerful regularizations can inspire novel models for different learning tasks, and, without much additional effort, their nonlinear extensions.

\section{OPLS} \label{sec:opls}
In this section, we will first briefly introduce OPLS for multivariate regression analysis, and then apply it to subspace learning for multi-class classification, and finally build its connections to LDA and CCA.

\subsection{Multivariate regression analysis} \label{sec:mvr}
OPLS \cite{worsley1997characterizing} is proposed  as a multivariate analysis method for feature extraction. Let $\{ (\mathbf{x}_i, \mathbf{y}_i)\}_{i=1}^n$ be a set of $n$ data pairs with input $\mathbf{x}_i \in \mathbb{R}^{d}$ and output $\mathbf{y}_i \in \mathbb{R}^{o}$. Denote the matrix representations of the $n$ data pairs as $X=[\mathbf{x}_1, \ldots,\mathbf{x}_n] \in \mathbb{R}^{d \times n}$ and $Y =[\mathbf{y}_1,\ldots,\mathbf{y}_n] \in \mathbb{R}^{o\times n}$. Let 
\begin{align}
\hat{X} = X H_n, \label{eq:X-center}
\end{align}
be the centered matrix of $X$, where the centering matrix $H_n = I_n - \frac{1}{n} \mathbf{1}_n \mathbf{1}_n$, $I_n$ is the identity matrix of size $n$ and $\mathbf{1}_n$ is the $n$-dimensional column vector of all ones. 

OPLS  \cite{worsley1997characterizing} aims to learn a projection matrix $P \in \mathbb{R}^{d \times k}$ to transform input data from a $d$-dimensional space $\mathbb{R}^d$ to a $k$-dimensional space $\mathbb{R}^k$ by solving
\begin{align}
\max_{P} \tr(P^{\T} \hat{X} \hat{Y}^{\T} \hat{Y} \hat{X}^{\T} P) ~:\st P^{\T} \hat{X} \hat{X}^{\T} P = I_k, \label{op:opls-max}
\end{align}
with the generalized eigenvalue decomposition \cite{golub2012matrix}, where $\hat{Y} = Y H_n$. Later, we will use (\ref{op:opls-max}) with a different $\hat{Y}$.

In \cite{roweis1999linear,arenas2008efficient}, problem (\ref{op:opls-max}) is reformulated as a multivariate regression problem in the mean square error sense: 
\begin{align}
\min_{P, W} \| \hat{Y} - W^{\T} P^{\T} \hat{X} \|_F^2, \label{op:opls-mse}
\end{align}
where coefficient matrix $W \in \mathbb{R}^{k \times o}$, and $\|\cdot\|_F$ is the Frobenius norm.

Problems (\ref{op:opls-max}) and (\ref{op:opls-mse}) are closely related. To see this, we  eliminate $W$ from (\ref{op:opls-mse}) as follows.
The first order optimality condition of  (\ref{op:opls-mse}) for $W$ is
\begin{align}
-2 P^{\T} \hat{X} ( \hat{Y} - W^{\T} P^{\T} \hat{X}  )^{\T} = 0.
\end{align}
Assuming that matrix $\hat{X} \hat{X}^{\T}$ is positive definite (this will be resolved later by introducing regularization), we have
\begin{align}
W = (P^{\T} \hat{X} \hat{X}^{\T} P)^{-1} P^{\T} \hat{X} \hat{Y}^{\T}. \label{eq:opls-W}
\end{align}
Substituting (\ref{eq:opls-W}) into (\ref{op:opls-mse}), we get a reformulated problem of  (\ref{op:opls-mse})
\begin{align}
\min_P \| \hat{Y} \|_F^2 - \tr( (P^{\T} \hat{X} \hat{X}^{\T} P)^{-1} P^{\T} \hat{X} \hat{Y}^{\T} \hat{Y} \hat{X}^{\T} P). \label{op:opls-p}
\end{align} 
Define $\hat{P} = P (P^{\T} \hat{X} \hat{X}^{\T} P)^{-1/2} $. Problem (\ref{op:opls-p}) is equivalent to 
\begin{align}
\max_ {\hat{P}} \tr(\hat{P}^{\T} \hat{X} \hat{Y}^{\T} \hat{Y} \hat{X}^{\T} \hat{P}) ~:\st~ \hat{P}^{\T} \hat{X} \hat{X}^{\T} \hat{P} = I_k, \label{op:opls-p-max}
\end{align}
which is the same as (\ref{op:opls-max}).  The two formulations (\ref{op:opls-max}) and (\ref{op:opls-p-max}) imply that an optimal solution $P_*$ of (\ref{op:opls-mse}) can always be transformed to an optimal solution $P $  of (\ref{op:opls-max}) via $P= P_* (P_*^{\T} \hat{X} \hat{X}^{\T} P_*)^{-1/2}$. On the other hand,
it is obvious that an optimal solution of (\ref{op:opls-max}) is also an optimal solution of (\ref{op:opls-p}). Hence, problems (\ref{op:opls-max}) and (\ref{op:opls-p-max}) are equivalent.


The least squares formulation (\ref{op:opls-mse}) of OPLS possesses many appealing properties.

1) Regularization is naturally made possible to prevent singularity issue in (\ref{op:opls-mse}). Regularized OPLS is formulated as
\begin{align}
\min_{P, W} \| \hat{Y} - W^{\T} P^{\T} \hat{X} \|_F^2 + \lambda \| PW \|_F^2, \label{op:ropls-mse}
\end{align}
where $\lambda>0$ is a regularization parameter. Correspondingly, the regularized OPLS (\ref{op:ropls-mse}) is equivalent to
\begin{align}
\!\!\min_P \| \hat{Y} \|_F^2 - \tr( (P^{\T} (\hat{X} \hat{X}^{\T} \!\!+\!\! \lambda I_d) P)^{-1} P^{\T} \hat{X} \hat{Y}^{\T} \hat{Y} \hat{X}^{\T} P), \label{op:ropls-p}
\end{align}
or, equivalently,
\begin{align}
\!\!\max_{P} \tr(P^{\T} \!\hat{X} \hat{Y}^{\T} \!\hat{Y} \hat{X}^{\T} P) :\st P^{\T}\!(\hat{X} \hat{X}^{\T} +\lambda I_d)P \!=\! I_k. \label{op:ropls-max}	
\end{align}
The singularity issue is resolved since $\hat{X} \hat{X}^{\T} + \lambda I_d$ is positive definite for any given $X$. 

2) OPLS also provides a built-in multivariate regression model for predicting an output of any given input $\mathbf{x}$  by
\begin{align}
f(\mathbf{x}) = W^{\T} P^{\T} (\mathbf{x} - \bm{ \mu}),
\end{align}
where $\bm{\mu} = \frac{1}{n} \sum_{i=1}^n \mathbf{x}_i$ is the mean of the input training data.

3) CCA and OPLS are equivalent, as proved in \cite{sun2009equivalence}.

\subsection{Multi-class classification} \label{sec:mcc}
OPLS can be applied to multi-class classification. Suppose that there are $c$ classes in data $\{(\mathbf{x}_i, y_i)\}_{i=1}^n$, where each class label $y_i  \in \{1, \ldots, c\}$.  Define $Y = [\mathbf{y}_1,\ldots,\mathbf{y}_n]\in\{0,1\}^{c\times n}$ by one-hot representation of class labels that transforms categorical class labels to multivariate outputs:  $Y_{r,i}=(\mathbf{y}_i)_r=1$ if the $i$th label $y_i = r$, and otherwise $0$s. $Y$ is the indicator matrix of data and possesses the following properties:
\begin{enumerate}
	\item the sum of each column of $Y$ is one:
	\begin{align}
	Y^{\T} \mathbf{1}_c = \mathbf{1}_n,
	\end{align}
	which means every data point belongs to one and only one class;
	
	\item the counting matrix $\Sigma$ of class labels 
	\begin{align}
	\Sigma= \begin{bmatrix}
	n_1&\\
	& n_2 \\
	& & \ddots\\
	& && n_c
	\end{bmatrix} = Y Y^{\T} \in \mathbb{R}^{c \times c},
	\end{align}
	where $n_r = \sum_{i=1}^n Y_{r,i} $ is the number of data points in class $r$;
	
	\item the similarity matrix $Q = Y^{\T} \Sigma^{-1} Y$ is both row and column normalized to $1$. In fact, $Q$ is symmetric and
	\begin{align}
	Q \mathbf{1}_n =  Y^{\T} \Sigma^{-1} \begin{bmatrix}
	n_1 \\
	n_2 \\
	\vdots\\
	n_c
	\end{bmatrix} = Y^{\T} \mathbf{1}_c = \mathbf{1}_n; \label{eq:Y-normalization}
	\end{align}
	
	\item the centered matrix of $Q$, denoted by $\hat{Q}$, is given by
	\begin{align}
	\hat{Q} =& H_n Q H_n \nonumber\\
	=& (I_n - \frac{1}{n} \mathbf{1}_n \mathbf{1}_n^{\T}) Q (I_n - \frac{1}{n} \mathbf{1}_n \mathbf{1}_n^{\T}) \nonumber \\
	=& Q -  \frac{1}{n} \mathbf{1}_n \mathbf{1}_n^{\T} Q- Q \frac{1}{n} \mathbf{1}_n \mathbf{1}_n^{\T} + \frac{1}{n^2 } \mathbf{1}_n \mathbf{1}_n^{\T} \mathbf{1}_n \mathbf{1}_n^{\T} \nonumber\\
	=&Q - \frac{1}{n} \mathbf{1}_n \mathbf{1}_n^{\T}. \label{eq:Q1}
	\end{align} 
\end{enumerate}

Below, we will show that LDA  for supervised classification is a special case of OPLS (\ref{op:opls-max})  with $\hat{X}$ in (\ref{eq:X-center}) and 
\begin{align}
\hat{Y} = \Sigma^{-1/2} Y. \label{con:lda}
\end{align}
We have the following equalities:
\begin{align}
\hat{X} \hat{X}^{\T} &= X H_n H_n X^{\T} = X H_n X^{\T} =: \hat{C}, \label{eq:cov}\\
\hat{X} \hat{Y}^{\T} \hat{Y} \hat{X}^{\T} &= X H_n Y^{\T} [\Sigma^{-\frac{1}{2}}]^{\T} \Sigma^{-\frac{1}{2}} Y H_n X^{\T} \nonumber\\
&=X H_n Y^{\T} \Sigma^{-1} Y H_n X^{\T}  \nonumber\\
&=X H_n Q H_n X^{\T} \nonumber\\
&= X (Q - \frac{1}{n} \mathbf{1}_n \mathbf{1}_n^{\T}) X^{\T} \nonumber\\
&=:  S_b. \label{eq:bcs}
\end{align}
It is clear that $\hat{C}$ is the covariance matrix of $X$ and $S_b$ is the between-class scatter matrix \cite{yan2005graph}. Accordingly, the within-class scatter matrix $S_w \equiv \hat{C}-S_b = X(I_n - Q) X^{\T}$. Hence, OPLS (\ref{op:opls-max}) with (\ref{con:lda}) reduces to LDA  \cite{yan2005graph}, that is,
\begin{align}
\max_P \tr(P^{\T} S_b P): \st P^{\T} \hat{C} P = I_k.
\end{align}
Hence, LDA has a close relationship with the least squares formulation (\ref{op:opls-mse}) of OPLS.  

For multi-class classification, the least squares formulation (\ref{op:opls-mse}) of OPLS can be a useful tool to reveal the relationships among different models. It has been explored in the past.
LDA has previously been formulated as a least squares problem for binary classification \cite{bishop2006pattern} and multi-class classification \cite{ye2007least}. In addition, CCA as a supervised method has been shown to be equivalent to LDA for multi-class classification \cite{hastie1995penalized}, where the class labels are treated as from one view and the input data points as from another.

The properties of OPLS for multivariate regression in Subsection \ref{sec:mvr} are naturally inherited by OPLS for multi-class classification. The special indicator output matrix contributes to additional useful properties. For example, the regularized OPLS (\ref{op:ropls-mse}) can recast the shrinkage estimator of the covariance matrix for LDA  \cite{yu2001direct}. 
According to (\ref{op:opls-mse}), the multi-class decision function can be formulated as 
\begin{align}
y = \min_{r=1,\ldots,c} \mathbf{w}_r^{\T} P^{\T} (\mathbf{x} - \bm{ \mu}), \label{eq:predictive-class}
\end{align}
where $W = [\mathbf{w}_1,\ldots,\mathbf{w}_c] \in \mathbb{R}^{k \times c}$ is the coefficient matrix of the multi-class classifier.

\section{Regularized Multi-view OPLS} \label{reg:rmvopls}
We propose to extend OPLS  to multi-view classification whose input data consists of multi-views. For the ease of reference, we name our proposed method as multi-view OPLS (MvOPLS). Next, we will first present the formulation of our MvOPLS and then explore its regularization power.

\subsection{Multi-view formulation}
Let $\{ (\mathbf{x}_i^{(1)}, \ldots, \mathbf{x}_i^{(v)}, y_i)\}_{i=1}^n$ be the labeled data consisting of $v$ views, where the $i$th inputs $\mathbf{x}_i^{(s)}\in \mathbb{R}^{d_s}$ of all views have class label $y_i \in \{1,\ldots, c\}$ of $c$ classes. Represent the $n$ data points of the $s$th view by matrix $X_s = [\mathbf{x}_1^{(s)}, \ldots, \mathbf{x}_n^{(s)}] \in \mathbb{R}^{d_s \times n}$.  Let $P_s \in \mathbb{R}^{d_s \times k}$ be the projection matrix to transform $\mathbf{x}_i^{(s)}$ from $\mathbb{R}^{d_s}$ to $\mathbf{z}^{(s)} = P_s^{\T} \mathbf{x}_i^{(s)}$ in the common space $\mathbb{R}^k$, and let $Z_s =[\mathbf{z}^{(s)}_1,\ldots,\mathbf{z}_n^{(s)}]$ $= P_s^{\T} X_s \in \mathbb{R}^{k \times n}$. The label matrix $Y$ is  as defined in Subsection \ref{sec:mcc} by one-hot representation.

Assume that all view classifiers share the same coefficient matrix $W$. MvOPLS is formulated as minimizing the sum of OPLS objectives for all $v$ views, given by
\begin{align}
\min_{\{P_s\},W}  \sum_{s=1}^v \| \widetilde{Y} - W^{\T} P_s^{\T} \widetilde{X}_s \|_F^2, \label{op:mvopls-ls}
\end{align}
where $\widetilde{Y} \in \mathbb{R}^{c \times n}$ and $\widetilde{X}_s \in \mathbb{R}^{d_s \times n}$ are matrices transformed from $Y$  and $X_s$, respectively, dependent of particular MvOPLS models. 
For example, the most natural choice is
$\widetilde{X}_s=\hat{X}_s:=X_sH_n$ for all $s$ and
$\widetilde{Y}=YH_n$. Later in Section \ref{sec:reformulation}, we will introduce various
other choices of $\widetilde{Y}$ and $\widetilde{X}_s$ and their
connections to existing methods.


For convenience of analysis, denote by $d=\sum_{s=1}^v d_s$ the total number of features from all $v$ views, and by
\begin{align}
P \!=\!\! \begin{bmatrix}
P_1\\
P_2\\
\vdots\\
P_v
\end{bmatrix}\!\!\! \in\! \mathbb{R}^{d \times k}, 
\widetilde{X} \!=\!\! \begin{bmatrix}
\widetilde{X}_1\\
\widetilde{X}_2\\
\vdots\\
\widetilde{X}_v
\end{bmatrix} \!\!\!\in\! \mathbb{R}^{d \times n},
X \!=\!\! \begin{bmatrix}
X_1\\
X_2\\
\vdots\\
X_v
\end{bmatrix} \!\!\!\in\! \mathbb{R}^{d \times n}, \label{eq:stackX}
\end{align}
the concatenations of $\{P_s\}$, $\{\widetilde{X}_s\}$, and $\{X_s\}$. Define
\begin{align}
\widetilde{C}&=\widetilde{X} \widetilde{X}^{\T} =\begin{bmatrix}
\widetilde{C}_{1,1} & \widetilde{C}_{1,2} & \ldots & \widetilde{C}_{1,v}\\
\widetilde{C}_{2,1} & \widetilde{C}_{2,2} & \ldots & \widetilde{C}_{2,v}\\
\vdots& \vdots& \ddots & \vdots\\
\widetilde{C}_{v,1} & \widetilde{C}_{v,2} & \ldots& \widetilde{C}_{v,v} 
\end{bmatrix} \in \mathbb{R}^{d \times d}, 
\end{align}
and its block diagonal part
\begin{align}
\widetilde{C}_{\rm diag}  &= 
\begin{bmatrix}
\widetilde{C}_{1,1} \\
& \widetilde{C}_{2,2} \\
&&\ddots\\
&&& \widetilde{C}_{v,v} 
\end{bmatrix} \in \mathbb{R}^{d \times d}, \label{eq:cs}
\end{align}
where $\widetilde{C}_{s,t} = \widetilde{X}_s \widetilde{X}_t^{\T}, \forall s, t=1,\ldots,v$. Similarly define $C$ and $C_{\rm diag}$ in terms of $X$.

Following the same derivation procedure presented  in Subsection \ref{sec:mvr}, we obtain the first order optimality condition of (\ref{op:mvopls-ls}) with respect to $W$, given by
\begin{align}
\sum_{s=1}^v   -2 P_s^{\T} \widetilde{X}_s ( \widetilde{Y} - W^{\T} P_s^{\T} \widetilde{X}_s  )^{\T} = 0,
\end{align}
and then the analytic solution of $W$ is 
\begin{align}
W &= \left(  \sum_{s=1}^v P_s^{\T} \widetilde{C}_{s,s} P_s \right)^{-1} \sum_{s=1}^v  P_s^{\T} \widetilde{X}_s \widetilde{Y}^{\T} \nonumber\\
&= ( P^{\T}  \widetilde{C}_{\rm diag} P )^{-1} P^{\T}  \widetilde{X} \widetilde{Y}^{\T}. \label{eq:mvopls-W}
\end{align}
Substituting the optimal $W$ in (\ref{eq:mvopls-W}) back into (\ref{op:mvopls-ls}), we obtain a reformulated problem of (\ref{op:mvopls-ls}) as
\begin{align}
\min_P \| \widetilde{Y} \|_F^2 - \tr( (P^{\T} \widetilde{C}_{\rm diag}  P)^{-1} P^{\T} \widetilde{X} \widetilde{Y}^{\T} \widetilde{Y} \widetilde{X}^{\T} P). \label{op:mvopls-p}
\end{align}
Or, equivalently,
\begin{align}
\max_P ~ \tr( P^{\T} \widetilde{X} \widetilde{Y}^{\T} \widetilde{Y} \widetilde{X}^{\T} P ) 
~:\st ~ P^{\T} \widetilde{C}_{\rm diag}  P = I_k.
\end{align}
It is worth noting that the objective
\begin{align}
\tr( P^{\T} \widetilde{X} \widetilde{Y}^{\T} \widetilde{Y} \widetilde{X}^{\T} P ) = \sum_{s=1}^v \sum_{t=1}^v \tr(P_s^{\T} \widetilde{X}_s \widetilde{Y}^{\T} \widetilde{Y} \widetilde{X}_t^{\T} P_t)
\end{align}
indicates that the optimal projections are the ones that maximizes the sum of $v^2$ cross-view OPLS objectives. This differs from the $v$ OPLS models on each individual view. Comparing with OPLS over a single concatenated $\widetilde{X}$, MvOPLS imposes its constraint on $\widetilde{C}_{\rm diag}$ instead of $\widetilde{C}$.

For the natural choice, $\widetilde{X}_s=\hat{X}_s := X_s H_n$, $\widetilde{X}=\hat{X} := X H_n$. Accordingly, $\widetilde{C}_{s,s} =\hat{C}_{s,s} := \hat{X}_s \hat{X}_s^{\T}$ is the covariance matrix of data points of the $s$th view. Moreover, $\widetilde{C}=\hat{C} := \hat{X} \hat{X}^{\T}$ is the covariance matrix of the concatenated data of all views and $\widetilde{C}_{\rm diag}=\hat{C}_{\rm diag}$ is its block diagonal part.

\subsection{Regularization} \label{sec:reg}
As MvOPLS (\ref{op:mvopls-ls}) is formulated as the least squares, regularization technique can be incorporated in order to regulate the model and integrate certain prior knowledge for multi-view learning. Due to the special structure, three types of regularizations can be added to MvOPLS with respect to different considerations of priors, including (i) model parameters, (ii) decision values, and (iii)  latent projected points. Some examples of the three types are shown below.

\subsubsection{Model parameters}
In (\ref{op:mvopls-ls}), model parameters include $W$ and $\{ P_s \}_{s=1}^v$. One can consider each product $P_s W$ as one single variable. The Tikhonov regularization is formulated to mitigate the problem of multicollinearity in linear regression in (\ref{op:ropls-mse}) for single-view learning. For multi-view learning, the weighted Tikhonov regularizer can be defined as
\begin{align}
\mathcal{R}_{\textrm{tikh}} (W, \{P_s\}) = \sum_{s=1}^v \gamma_s \| P_s W \|_F^2 = \tr( W^{\T} P^{\T} \Gamma P W ), \label{eq:r-tikh}
\end{align}
where $\gamma_s \geq 0$ is the weight for view $s$ and the block diagonal matrix
\begin{align}
\Gamma &= 
\begin{bmatrix}
\gamma_1 I_{d_1} \\
& \gamma_2 I_{d_2} \\
&&\ddots\\
&&& \gamma_v I_{d_v}
\end{bmatrix} \in \mathbb{R}^{d \times d}. 
\end{align}

\subsubsection{Decision values of multi-class classifiers}
According to  (\ref{op:mvopls-ls}), the decision function of the multi-class classifier for the $s$th view can be written as 
\begin{align}
g_s({\mathbf{x}}^{(s)}) = W^{\T} P_s^{\T} {\mathbf{x}}^{(s)}, 
\end{align}
where ${\mathbf{x}}^{(s)}$ is an input data point of the $s$th view. Since each view classifier generates decision values for the same set of instances, it is proper to constrain these decision values. 

One might require the decision values of the mean vectors of views are close to each other across all views. The decision value of the mean vector of the $s$th view is
\begin{align}
\overline{g}_s = \frac{1}{n} W^{\T} P_s^{\T} {X}_s \mathbf{1}_n, \forall s=1,\ldots,v.
\end{align}
To impose their closeness, it is natural to minimize the following quantity
\begin{align}
 \mathcal{R}_{\textrm{mean}} (W, \{P_s\}; \{X_s\}) = & \frac{n}{2v}\sum_{s=1}^v \sum_{t=1}^v \| \overline{g}_s  - \overline{g}_t \|^2 \nonumber\\
=& \frac{1}{n} \sum_{s=1}^v \tr(W^{\T} P_s^{\T} {X}_s \mathbf{1}_n \mathbf{1}_n^{\T} {X}_s^{\T} P_s W) \nonumber\\
&- \frac{1}{n v} \tr(W^{\T} P^{\T} {X} \mathbf{1}_n \mathbf{1}_n^{\T} {X} ^{\T} P W). \label{eq:r-mean}
\end{align}

In addition to imposing consistency among the decision values of view centers, the decision values can be parameterized by the representer theorem \cite{scholkopf2002learning} for least squares, that is, for view $s$, the decision values can be represented by a weighted combination of the input data points, given by
\begin{align}
P_s W &= {X}_s \bm{\beta}_s, \label{eq:rep-th}\\
g_s({X}_s) &= W^{\T} P_s^{\T} {X}_s = \bm{\beta}_s^{\T} {X}_s^{\T} {X}_s,
\end{align}
where $\bm{\beta}_s \in \mathbb{R}^{n \times c}$ is the weight vector for the $s$th view. Note that ${X}_s^{\T} {X}_s$ is the linear kernel of the $s$th view. 
According to (\ref{eq:rep-th}), we can obtain $\bm{\beta}_s = {X}_s^{\dagger} P_s W $ where ${X}_s^{\dagger} = ( {X}_s^{\T}  {X}_s)^{-1}  {X}_s^{\T}$ is the pseudo-inverse of $ {X}_s$.
If all views have similar kernels, it is reasonable to assume that the weight vectors $\{\bm{\beta}_s\}_{s=1}^v$ are close to each other, i.e., to minimize
\begin{align}
\mathcal{R}_{\beta} (W, \{P_s\}; \{X_s\}) &= \frac{1}{2} \sum_{s=1}^v \sum_{t=1}^v \| \bm{\beta}_s  - \bm{\beta}_t \|_F^2 \nonumber \\
&= \tr(W^{\T} P^{\T} M P W), \label{eq:r-beta}
\end{align} 
where $M$ is a block matrix with the $(s,t)$th block 
\begin{align}  \label{eq:r-beta-M}
M_{s,t} = \left\{
\begin{array}{ll}
(v-1) ({X}_s^{\dagger})^{\T} {X}_s^{\dagger}, & s=t ,\\
- ({X}_s^{\dagger})^{\T} {X}_{t}^{\dagger}, & s\not=t.
\end{array}
\right.
\end{align}

Alignment between the similarity matrix of class labels and the predicted values can be useful criterion for learning projections, too. Denote by $g_s({X}_s) = W^{\T} P_s^{\T} {X}_s$ the predicted soft labels. The HSIC criterion \cite{gretton2005measuring} for multi-view data can be used as a regularizer
\begin{align}
&\mathcal{R}_{\rm hsic} (W, \{P_s\}; \{X_s\}, Y) \nonumber\\
=& - \sum_{s=1}^v \tr( g_s({X}_s)^{\T}  g_s({X}_s) H_n{Y}^{\T} \Sigma^{-1} {Y} H_n) \nonumber \\
=& - \sum_{s=1}^v \tr( W^{\T} P_s^{\T} {X}_s H_n {Y}^{\T} \Sigma^{-1} {Y} H_n {X}_s ^{\T} P_s W ). \label{eq:r-hsic}
\end{align}

It is clear that (\ref{eq:r-mean})  captures the first order statistics of the decision values, while (\ref{eq:r-beta}) and (\ref{eq:r-hsic}) characterize the second order statistics.

\subsubsection{Projected points onto the common space}
The projected data in the common space is given by
\begin{align}
\widetilde{Z}_s = P_s^{\T} \widetilde{X}_s, \forall s=1,\ldots,v.
\end{align}
It is reasonable to expect that the projected points of the same instance from different views are close. In this case, the view-specific classifiers should perform similarly over all views. This hypothesis can be formulated by minimizing the following regularizer
\begin{align}
\mathcal{R}_{\textrm{cca}} (\{P_s\}; \{\widetilde{X}_s\}) &= \frac{1}{2} \sum_{s=1}^v \sum_{t=1}^v \| \widetilde{Z}_s - \widetilde{Z}_t \|_F^2 \nonumber\\
&= v \tr( P^{\T} \widetilde{C}_{\rm diag} P) - \tr(P^{\T} \widetilde{C} P).
\label{eq:r-cca}
\end{align}
Note that (\ref{eq:r-cca}) has been explored to generalize CCA, based on pairwise distances between projected points of different views \cite{via2007learning}.

LDA can also be used to regulate the projection of each view. It can be written as
\begin{align}
\mathcal{R}_{\rm lda}(\{P_s\}; \{X_s\}) = \sum_{s=1}^v \tr( P_s^{\T} X_s R X_s^{\T} P_s ), \label{eq:r-hsic-p}
\end{align}
where $R =H_n -\lambda(Q - \frac{1}{n} \mathbf{1}_n \mathbf{1}_n^{\T} ), \forall s=1,\ldots,v$ since $X_s R X_s^{\T}$ is the difference between covariance matrix $\hat{C}_{s,s}$ and the scaled between-class scatter matrix $\lambda {X}_s  (Q - \frac{1}{n} \mathbf{1}_n \mathbf{1}_n^{\T} ) {X}_s^{\T}$, and the scaling parameter $\lambda$ is the tradeoff parameter between the two scatter matrices, so (\ref{eq:r-hsic-p}) has the functionality analogous to the fractional formulation of LDA for each view.

For regularizations over projected points, it is proper to add certain weighting constraints for the ease of optimization. Instead of (\ref{eq:r-cca}) and (\ref{eq:r-hsic-p}), more generally we use
\begin{align}
\mathcal{R}_{\textrm{cca}} (\{P_s\}; \{\widetilde{X}_s\}, \Omega) &=  \tr( \Omega^{-1} P^{\T}  ( v \widetilde{C}_{\rm diag} - \widetilde{C}) P), \label{eq:r-cca-o}\\
\mathcal{R}_{\rm lda}(\{P_s\}; \{X_s\}, Y, \Omega) &= \sum_{s=1}^v \tr( \Omega^{-1} P_s^{\T} {X}_s R {X}_s^{\T}  P_s ),  \label{eq:r-hsic-p-o}
\end{align}
where $\Omega \in \mathbb{R}^{k \times k}$ is symmetric positive definite. It is clear that $\Omega = I_k$ yields the ones in (\ref{eq:r-cca}) and (\ref{eq:r-hsic-p}).

\subsection{A unified framework}
Abstracting regularizations (\ref{eq:r-tikh}), (\ref{eq:r-mean}), (\ref{eq:r-beta}), (\ref{eq:r-hsic}), (\ref{eq:r-cca}), (\ref{eq:r-hsic-p}), (\ref{eq:r-cca-o}) and (\ref{eq:r-hsic-p-o}), we formulate a unified regularized MvOPLS framework given by
\begin{align}
\min_{\{P_s\},W} &  \sum_{s=1}^v \| \widetilde{Y} - W^{\T} P_s^{\T} \widetilde{X}_s \|_F^2 + \tr(W^{\T}P^{\T} A P W) + \tr( \Omega^{-1} P^{\T} B P), \label{op:ropls-ls}
\end{align}
where $A$ and $B$ are some specific matrices derived from regularizations such as ones in Subsection \ref{sec:reg}.  

Special choices of $\Omega$ may simplify (\ref{op:ropls-ls}) for ease of its numerical treatment. For example, for $\Omega = P^{\T}(\widetilde{C}_{\rm diag} + A)P$,  (\ref{op:ropls-ls}) is equivalent to
\begin{align}
\max_P &~ \tr( P^{\T} (\widetilde{X} \widetilde{Y}^{\T} \widetilde{Y} \widetilde{X}^{\T} -  B) P ) \label{op:ropls-ged}\\
\st &~ P^{\T} (\widetilde{C}_{\rm diag} +  A) P = I_k, \nonumber
\end{align}
in such a way that an optimizer $P$ of (\ref{op:ropls-ged})  yields an optimizer $(P, W)$ of (\ref{op:ropls-ls}) with
\begin{align}
W =  (P^{\T} (\widetilde{C}_{\rm diag} + A) P ) ^{-1} P^{\T}  \widetilde{X} \widetilde{Y}^{\T} = P^{\T}  \widetilde{X} \widetilde{Y}^{\T}. \label{eq:M}
\end{align}
It is worth noting that the unified framework (\ref{op:ropls-ged}) can be solved by the generalized eigenvalue solver \cite{golub2012matrix}. 

\section{Examples of regularized MvOPLS} \label{sec:reformulation}
We will show various instantiations of the proposed regularized OPLS framework (\ref{op:ropls-ls}) by reformulating existing methods in the literature, including MCCA \cite{nielsen2002multiset,via2007learning}, MvLDA \cite{kan2016multi},  MvDA \cite{kan2015multi}, MvDA-VC \cite{kan2015multi}, MvMDA \cite{cao2017generalized}, MLDA \cite{sun2015multiview}  and GMA  \cite{sharma2012generalized}, and then, we discuss the differences among these methods and further propose some novel formulations.


\subsection{Reformulations of existing methods} \label{sec:examples}
\subsubsection{MCCA} \label{sec:mcca}
MCCA \cite{nielsen2002multiset,via2007learning} is an unsupervised subspace learning method. As the data label is unknown, the simplest approach is to assume that each instance is in its own class, so $c=n$. By simply assigning a unique class label to each instance, the unlabeled data is transformed to labeled data with $Y = I_n$. MvOPLS (\ref{op:mvopls-ls}) with 
$\widetilde{X} = \hat{X} = XH_n$, $\widetilde{Y} = I_n$, and regularizer $\mathcal{R}_{\rm tikh}$ with all $\gamma_s=\gamma$ is instantiated as
\begin{align}
\min_{\{P_s\},W} &   \sum_{s=1}^v \| \widetilde{Y} - W^{\T} P_s^{\T} \widetilde{X}_s \|_F^2 +  \mathcal{R}_{\textrm{\rm tikh}} (W, \{P_s\}).
\end{align}
It is equivalent to the regularized MCCA formulation
\begin{align}
\max_P & \tr(P^{\T} X H_n X^{\T} P) := \sum_{s=1}^v\sum_{t=1}^v \tr(P^{\T} \hat{C}_{s,t} P)\\
\st & \sum_{s=1}^ v P_s^{\T} (\hat{C}_{s,s} + \gamma I_{d_s}) P_s = I_k, \nonumber
\end{align}
where $\hat{C}_{s,t} = X_s H_n X_t^{\T}, \forall s, t=1,\ldots,v$.

\subsubsection{MvLDA}
MvLDA \cite{kan2016multi} is equivalent to LDA on $X$ of (\ref{eq:stackX}). In other words, it degrades to OPLS. MvOPLS (\ref{op:mvopls-ls})  with $\widetilde{X} = \hat{X} = XH_n$, $\widetilde{Y}=\Sigma^{-1/2} Y$ and regularizer $\mathcal{R}_{\rm tikh}$ with all $\gamma_s = \gamma$ is formulated as 
\begin{align}
\min_{\{P_s\},W} &   \| \widetilde{Y} - W^{\T} P^{\T} \widetilde{X} \|_F^2 + \mathcal{R}_{\textrm{tikh}} (W, \{P_s\}),
\end{align}
which leads to the following optimization problem 
\begin{align}
\max_P & \tr(P^{\T} S_b P) \\
\st &~ P^{\T} (X H_n X^{\T} + \gamma I_d) P = I_k \nonumber,
\end{align}
where $S_b = X (Q - \frac{1}{n} \mathbf{1}_n \mathbf{1}_n^{\T}) X^{\T}$ is the between-class scatter matrix as discussed in Subsection \ref{sec:mcc}.

\subsubsection{MvDA}
MvDA \cite{kan2016multi} was originally motivated in terms of some specific definitions of within-class scatter and between-class scatter matrices for multi-view data. It can be simply realized as MvOPLS (\ref{op:mvopls-ls})  with $\widetilde{X}= \hat{X} = XH_n$, $\widetilde{Y}=\Sigma^{-1/2} Y$ and regularizer $\mathcal{R}_{\rm mean}$, that is,
\begin{align}
\min_{\{P_s\},W} &   \sum_{s=1}^v \| \widetilde{Y} \!-\! W^{\T}\!P_s^{\T} \widetilde{X}_s \|_F^2 \!+\! \mathcal{R}_{\textrm{mean}} (W, \{P_s\}; \{X_s\}). \label{op:mvda-ls}
\end{align}
Accordingly, we have 
\begin{align*}
& \tr(P^{\T} \widetilde{C}_{\rm diag}   P)+  \frac{1}{n} \sum_{s=1}^v \tr(P_s^{\T} {X}_s \mathbf{1}_n \mathbf{1}_n^{\T} {X}_s^{\T} P_s)   \\
=& \sum_{s=1}^v \tr(P_s^{\T} {X}_s (H_n + \frac{1}{n} \mathbf{1}_n \mathbf{1}_n^{\T}) {X}_s^{\T} P_s)\\
=& \sum_{s=1}^v \tr(P_s^{\T} {X}_s {X}_s^{\T} P_s) \\
=& P^{\T} C_{\rm diag}  P,
\end{align*}
and, by using (\ref{eq:Q1}), 
\begin{align*}
\tr( P^{\T} \widetilde{X} \widetilde{Y}^{\T} \widetilde{Y} \widetilde{X}^{\T} P ) =&\tr(  P^{\T} X H_n Y^{\T} \Sigma^{-1} Y H_n X   P)\\
=& \tr(  P^{\T} X H_n Q H_n X   P)\\
=& \tr(P^{\T} X (Q - \frac{1}{n} \mathbf{1}_n \mathbf{1}_n^{\T}) X   P).
\end{align*}
Problem (\ref{op:mvda-ls}) is equivalent to
\begin{align}
\max_P &~ \tr(P^{\T} X (Q - \frac{1}{n} \mathbf{1}_n \mathbf{1}_n^{\T}) X   P) \label{op:mvda}\\
\textrm{s.t.} &~ P^{\T}  (C_{\rm diag} - \frac{1}{nv} X \mathbf{1}_n \mathbf{1}_n^{\T} X^{\T}) P = I_k.  \nonumber
\end{align}
Problem (\ref{op:mvda}), although not the same as MvDA, recovers the projection matrices obtained by MvDA. 
In addition, regularizer $\mathcal{R}_{\rm tikh}$ can also be applied to prevent possible ill-posedness in (\ref{op:mvda}), i.e., $C_{\rm diag} - \frac{1}{nv} X \mathbf{1}_n \mathbf{1}_n^{\T} X^{\T}$ being singular.

\subsubsection{MvDA-VC}
MvDA with view consistency (MvDA-VC) can be formulated as regularized MvOPLS  (\ref{op:mvda-ls})  also imposed with regularizer $\mathcal{R}_{\beta}$
\begin{align}
\min_{\{P_s\},W} &   \sum_{s=1}^v \| \widetilde{Y} - W^{\T} P_s^{\T} \widetilde{X}_s \|_F^2 \nonumber \\
&\!\!\!\!\!\!\!\!\!\!+ \mathcal{R}_{\textrm{mean}} (W, \{P_s\}; \{X_s\}) + \lambda \mathcal{R}_{\beta} (W, \{P_s\}; \{X_s\}). \label{op:mvda-vc-ls}
\end{align}
It is equivalent to
\begin{align}
\max_P &~ \tr(P^{\T} X (Q - \frac{1}{n} \mathbf{1}_n \mathbf{1}_n^{\T}) X   P) \label{op:mvda-vc}\\
\textrm{s.t.} &~ P^{\T}  (C_{\rm diag} - \frac{1}{nv} X \mathbf{1}_n \mathbf{1}_n^{\T} X^{\T} + \lambda M) P = I_k,  \nonumber
\end{align}
where $M$ is defined in (\ref{eq:r-beta-M}).  Similarly, 
It can be verified that problem (\ref{op:mvda-vc}) can recover the projections obtained by MvDA-VC \cite{kan2016multi}.

\subsubsection{MvMDA} \label{sec:MvMDA}
MvMDA \cite{cao2017generalized} is proposed to maximize the distances between different class centers across different views and minimize the within-class scatter. It can be reformulated as MvOPLS (\ref{op:mvopls-ls}) with $\widetilde{X} = \hat{X}= X$, $\widetilde{Y}=  H_c \Sigma^{-1} Y$ and regularizer $\mathcal{R}_{\rm hsic}$, given by
\begin{align}
\min_{\{P_s\},W} &   \sum_{s=1}^v \| \widetilde{Y} \!-\! W^{\T} \!P_s^{\T} \widetilde{X}_s \|_F^2 \!+\! \mathcal{R}_{\textrm{hsic}} (\!W, \!\{P_s\}; \!\{X_s\},\!Y\!). \label{op:mvmda-ls}
\end{align}
Accordingly, we have
\begin{align*}
\tr( P^{\T} \widetilde{X} \widetilde{Y}^{\T} \widetilde{Y} \widetilde{X}^{\T} P) 
=& \tr( P^{\T} \!X Y^{\T}  \Sigma^{-1} H_c^{\T} H_c   \Sigma^{-1} Y X^{\T}\! P ) \\
=& \sum_{s=1}^v \sum_{t=1}^v \tr(P_s^{\T} X_s  L_b X_s^{\T} P_s),
\end{align*}
where 
\begin{align*}
L_b &= Y^{\T} \Sigma^{-1} H_c   \Sigma^{-1} Y \\
&= \sum_{p=1}^c \frac{1}{n_p^2} \mathbf{u}_p \mathbf{u}_p^{\T} -  \frac{1}{c}\sum_{p=1}^c \sum_{q=1}^c \frac{1}{n_p n_q} \mathbf{u}_p \mathbf{u}_q^{\T}  \\
&= \frac{1}{c}\sum_{p=1}^c \sum_{q=1}^c   \left[ \frac{1}{n_p^2} \mathbf{u}_p \mathbf{u}_p^{\T}  - \frac{1}{n_p n_q} \mathbf{u}_p \mathbf{u}_q^{\T}  \right],
\end{align*}
and $\mathbf{u}_p^{\T}$ is the $p$th row of $Y$, i.e., $Y = [\mathbf{u}_1, \ldots,\mathbf{u}_c]^{\T}$. Moreover, we have
\begin{align*}
&\tr(P^{\T} C_{\rm diag} P) - \sum_{s=1}^v \tr(P_s^{\T} X_s Y^T \Sigma^{-1} Y X_s^{\T} P_s^{\T}) \\
=& \sum_{s=1}^v \tr(P_s^{\T}  X_s (I_n - Q) X_s^{\T} P_s)
\end{align*}
which is the sum of within-class scatter matrices of all views. By combining the above two terms, (\ref{op:mvmda-ls}) reduces equivalently to MvMDA formulation
\begin{align}
\max_P & \sum_{s=1}^v \sum_{t=1}^v \tr(P_s^{\T} X_s L_b X_s^{\T} P_s) \\
\st & \sum_{s=1}^v \tr(P_s^{\T}  X_s (I_n - Q) X_s^{\T} P_s) = I_k. \nonumber
\end{align}
Hence, MvMDA takes the centered label matrix as the output label of the least squares. The HSIC regularization replaces the covariance matrix of each view with within-class scatter matrix. 

\subsubsection{MLDA} \label{sec:MLDA}
MLDA \cite{sun2015multiview} is proposed to learn a common space such that the correlation between two views and the discrimination of each view can be maximized simultaneously. As shown in Subsubsection \ref{sec:mcca}, MCCA is a special case of MvOPLS (\ref{op:mvopls-ls}). Combined with regularizer $\mathcal{R}$, MvOPLS (\ref{op:mvopls-ls}) with $\widetilde{X} = XH_n$ and $\widetilde{Y} = I_n$ gives
\begin{align}
\min_{\{P_s\},W} &  \! \sum_{s=1}^v \| \widetilde{Y} \!-\! W^{\T} \!P_s^{\T} \!\widetilde{X}_s \|_F^2 \!+\! \mathcal{R}_{\rm lda}( \{P_s\};\!\{X_s\}, \!Y, \!\Omega),
\end{align}
where $\Omega=P^{\T} \hat{C}_{\rm diag} P$.
It is the same as
\begin{align}
\max_P & \tr( P^{\T} S P) ~:\st  P^{\T} \hat{C}_{\rm diag} P = I_k, \label{op:mlda}
\end{align}
where the matrix $S$ is a block matrix with the $(s,t)$th block given by
\begin{align}
S_{s,t} = \left\{
\begin{array}{ll}
\lambda X_s (Q - \frac{1}{n} \mathbf{1}_n \mathbf{1}_n^{\T} )X_s ^{\T} &  s=t,\\
\hat{C}_{s,t} & \textrm{otherwise,}
\end{array}
\right.
\end{align}
$\hat{C}_{s,s} - X_s R_s X_s^{\T} =\hat{C}_{s,s} - X_s(H_n -\lambda(Q - \frac{1}{n} \mathbf{1}_n \mathbf{1}_n^{\T} ))X_s^{\T} =\lambda  X_s (Q - \frac{1}{n} \mathbf{1}_n \mathbf{1}_n^{\T} )X_s ^{\T}$ is the between-class scatter matrix with scaling $\lambda$. Problem (\ref{op:mlda}) is the same as the formulation of MLDA \cite{sun2015multiview}.

\subsubsection{GMA}
GMA \cite{sharma2012generalized} is different from MLDA only on constraints. The within-class scatter is used in GMA instead of the total scatter used in MLDA.  MLDA can be transformed into GMA by adding regularization $\mathcal{R}_{\rm hsic}$, that is,
\begin{align}
&\!\!\!\!\!\!\!\!\!\!\min_{\{P_s\},W}    \sum_{s=1}^v \| \widetilde{Y} - W^{\T} P_s^{\T} \widetilde{X}_s \|_F^2 \nonumber\\
+& \mathcal{R}_{\rm lda}(\{P_s\};\{X_s\}, Y, \Omega) + \mathcal{R}_{\rm hsic} (W, \{P_s\};\{X_s\}, Y), \label{op:gma-mvopls}
\end{align}
with $\widetilde{X} = XH_n$, $\widetilde{Y} = I_n$ and $\Omega=\sum_{s=1}^v \tr(P_s^{\T}  X_s (I_n - Q) X_s^{\T} P_s)$.
Accordingly, the reformulated problem is
\begin{align}
\max_P & \tr( P^{\T} S P) \label{op:gma}\\
\st & \sum_{s=1}^v \tr(P_s^{\T}  X_s (I_n - Q) X_s^{\T} P_s) = I_k. \nonumber
\end{align}
Hence, the proposed model (\ref{op:gma-mvopls}) reduces to GMA.

\subsection{Discussions and new variants}
The seven existing methods discussed in Subsection \ref{sec:examples} can be partitioned into two categories based on whether MvOPLS takes class labels into account or not. The first category includes MCCA, MLDA and GMA, which assume that the unlabeled data is used to construct the least squares, i.e., unsupervised MvOPLS, while MLDA and GMA regulate MCCA by incorporating labeled data into the regularization terms, such as $ \mathcal{R}_{\rm lda}$ and $\mathcal{R}_{\rm hsic}$. In contrast, the second category consists of MvLDA, MvDA, MvDA-VC and MvMDA, which take supervised MvOPLS with labeled data into account together with unsupervised/supervised regularization terms. Among the methods in each category, the choices of both the input/output transformations and regularization terms become the key factors to distinguish one from another. For example, GMA differs from MLDA in that GMA takes additional regularizer $\mathcal{R}_{\rm hsic}$ to minimize the within-class scatter of each view. MvMDA takes the centered normalized label matrix and supervised MvOPLS with $\mathcal{R}_{\textrm{hsic}}$, while MvDA takes the normalized   label matrix and unsupervised MvOPLS with $\mathcal{R}_{\textrm{mean}}$.

With the guidance of the proposed unified framework (\ref{op:ropls-ls}) of the regularized MvOPLS, it is simple to design a new model by incorporating different inputs/outputs and regularizations. To demonstrate this point, we combine MvDA with $\mathcal{R}_{\rm cca}$ so that the projected points of each instance are forced to be close among all views for MvDA. This new variant can be formulated as
\begin{align}
&\!\!\!\!\!\!\!\!\!\!\min_{\{P_s\},W}    \sum_{s=1}^v \| \widetilde{Y} - W^{\T} P_s^{\T} \widetilde{X}_s \|_F^2 \nonumber\\
+& \mathcal{R}_{\rm mean}(W, \{P_s\};\{X_s\}) + \lambda \mathcal{R}_{\rm cca}(\{P_s\};\{X_s\}, \Omega),
\end{align}
where $\Omega = P^{\T}  (C_{\rm diag} - \frac{1}{nv} X \mathbf{1}_n \mathbf{1}_n^{\T} X^{\T}) P$.
Or, equivalently,
\begin{align}
\max_P &~ \tr\left(\!P^{\T}\! \left[X (Q - \frac{1}{n} \mathbf{1}_n \mathbf{1}_n^{\T}) X  + \lambda(\hat{C} - v \hat{C}_{\rm diag})\! \right] \! P\right) \label{op:mvda-cca}\\
\textrm{s.t.} &~ P^{\T}  (C_{\rm diag} - \frac{1}{nv} X \mathbf{1}_n \mathbf{1}_n^{\T} X^{\T}) P = I_k.  \nonumber
\end{align}
Due to the numerous ways of combinations, we in this paper will not enumerate all possible variants. For the task of interest, we recommend to use or design proper regularization terms to integrate into the proposed regularized MvOPLS framework (\ref{op:ropls-ls}), and by equivalence, (\ref{op:ropls-ged}).

\section{Deep Regularized MvOPLS} \label{sec:deep}
Regularized MvOPLS and its variants aim to learn a set of linear projections. 
Lately, extending a linear projection method to a nonlinear one via
the kernel trick becomes almost mechanical and immediate because it
is often very much straightforward. The case for MvOPLS methods so
far is no different. However, as pointed out in \cite{andrew2013deep}, kernel-based nonlinear extensions encounter several drawbacks:
\begin{enumerate}
	\item nonlinear representations are limited by the fixed kernel function;
	\item inner products between two of input data points are required, and so the training set has to be stored during the entire testing phrase;
	\item the time required to train a subspace learning model or compute the representations of new data points scales poorly with the size of the training set.
\end{enumerate}

To overcome these  drawbacks, the deep learning technique has been introduced to learn a set of nonlinear parametric functions for subspace-based multi-view learning \cite{andrew2013deep,kan2016multi,cao2017generalized}. For example, MvLDA and MvMDA use deep networks to learn nonlinear projections. MvLDA \cite{kan2016multi} takes a ratio trace formulation of LDA  as the objective function by concatenating all views, while MvMDA \cite{cao2017generalized} takes a trace ratio as the original objective function but minimizes it approximately by the generalized eigenvalue decomposition because of the availablity of numerical linear algebra packages. As discussed  in \cite{wang2007trace}, ratio trace and trace ratio actually yield  two different projections. 

Since regularized MvOPLS is clearly formulated as the ratio trace, or the generalized eigenvalue problem, we propose a nonlinear extension as the general problem 
\begin{align}
\max_P & \tr(P^{\T} A P) ~\st~ P^{\T} B  P = I_k, \label{op:ged}
\end{align}
where  $A=f(\{h_s(X_s)\}_{s=1}^v, Y)$ and $B=g(\{h_s(X_s)\}_{s=1}^v, Y)$ are some matrix-valued functions of $\{h_s(X_s)\}_{s=1}^v$ parameterized by $v$ independent deep networks  and the label matrix $Y$. For example, corresponding to each of the methods  in Section \ref{sec:reformulation}, $A$ and $B$ are as given there. 

Following \cite{andrew2013deep}, we will use multiple stacked layers with nonlinear activation functions as the deep network architecture. The $i$th layer in the network for the $s$th view has $m_s^i$ units, and the output layer has $k$ units. The output of the first layer for input $\mathbf{x}^{(s)}$ from the $s$th view is $h_s^1=\sigma(V^1_s \mathbf{x}^{(s)} +b^1_s) \in \mathbb{R}^{m_s^1}$, where $V^1_s \in \mathbb{R}^{m_s^1 \times d_s}$ is the weight matrix, $b^1_s \in \mathbb{R}^{m_s^1}$ is the vector of biases, and $\sigma: \mathbb{R} \rightarrow \mathbb{R}$ is a nonlinear activation function.
The output $h_s^1$ can then be used as the input to the next layer whose output $h_s^2 = \sigma(V^2_s h_s^1 +b^2_s) \in \mathbb{R}^{m_s^2}$, and the construction repeats $\ell$ times until the final output $h_s(\mathbf{x}_s) \equiv h_s^\ell= \sigma(V^\ell_s h_s^{\ell-1} +b^\ell_s) \in \mathbb{R}^{k}$ is reached. The same construction process can be used for each of the $v$ views. As a result, we have a set of nonlinear functions $\{h_s\}_{s=1}^v$ with $\ell$  layers  and their associated parameters $\{ V^i_s, b^i_s \}, \forall s=1,\ldots,v, i=1,\ldots,\ell$. To simplify the notation, we assume the nonlinear transformed matrix $h_s(X_s) \in \mathbb{R}^{k \times n}$ implicitly associates with its network parameters for the input $X_s$.

We further rewrite (\ref{op:ged}) as a standard eigenvalue problem, so that the gradient of the transformed objective with respect to network parameters can be computed by automatic differentiation tools. Specifically, let the Cholesky decomposition of $B$ be
\begin{align}
B= \Psi^{\T} \Psi.
\end{align}
To  ensure that $B$ is positive definite, the regularizer $\mathcal{R}_{\rm tikh}$ is applied to all the methods studied in this paper. Denote
\begin{align}
U = \Psi P \Rightarrow P = \Psi^{-1} U. \label{eq:convert}
\end{align}
Problem (\ref{op:ged}) is rewritten as
\begin{align}
\max_U &~ \tr( U^{\T} \Psi^{-\T}  A \Psi^{-1} U) ~:\st ~ U^{\T} U = I_k, \label{op:opls-eig}
\end{align}
which is equivalent to solving the eigen-decomposition
\begin{align}
\Psi^{-\T}  A \Psi^{-1}  U =  U \Lambda,
\end{align}
where $U$ and $\Lambda = \textrm{diag }(\lambda_1,\ldots,\lambda_d)$ are the eigenvector and eigenvalue matrices of $\Psi^{-\T}  A \Psi^{-1}$ with $\lambda_1\geq \lambda_2\geq \ldots \geq \lambda_d$. The optimal $U$ of (\ref{op:opls-eig}) consists of the eigenvectors corresponding to the top $k$ eigenvalues. After the optimal $U$ is obtained, we recover $P$ using (\ref{eq:convert}). The optimal objective function value is
\begin{align}
\tr(P^{\T} A P)  = \tr( U^{\T} \Psi^{-\T}  A \Psi^{-1} U ) = \sum_{m=1}^k \lambda_m. \label{eq:drmvopls}
\end{align}
Treating the negative of (\ref{eq:drmvopls}) as loss, we actually optimize the loss over network parameters $\{ V^i_s, b^i_s \}, \forall s=1,\ldots,v, i=1,\ldots,\ell$ and the projection matrix $\{ P_s\}_{s=1}^v$ simultaneously via the gradient descent method.

\section{Experiments} \label{sec:experiments}

\subsection{Data sets}
The statistics of the nine data sets with their corresponding descriptions are shown in Table \ref{tab:data}.

\begin{landscape}
	\begin{small}
		\begin{table*}
			\caption{Multi-view data sets used in the experiments, where the number of features for each view is shown inside the bracket.} \label{tab:data} 
			\centering
			\begin{tabular}{@{}crrcccccc@{}}
				\hline
				Data set & $n$ & c& view 1 &view 2 &view 3 &view 4 &view 5 &view 6 \\
				\hline
				Mfeat & 2000 & 10 &fac (216) & fou (76) & kar (64) & mor (6) & pix (240) & zer (47) \\
				Ads & 3279 & 2 & url+alt+caption (588) & origurl (495) & ancurl (472) & - & - & -\\
				Caltech101-7 & 1474 & 7 & CENTRIST (254) & GIST (512) & LBP (1180) & HOG (1008) & CH (64) & SIFT-SPM (1000)\\
				Caltech101-20 & 2386 & 20 & CENTRIST (254) & GIST (512) & LBP (1180) & HOG (1008) & CH (64) & SIFT-SPM (1000)\\
				Scene15 & 4310 & 15 & CENTRIST (254) & GIST (512) & LBP (531)  & HOG (360) & SIFT-SPM (1000) & -\\
				NUS-wide-object & 23953 & 31 & BOW (500) & CH (64) & CM55 (255) & CORR (144) & EDH (73) & WT (128)\\\hline
				Pascal & 1000& 20 & Text (100) & Image (1024)&-&-&-&-\\
				TVGraz & 2058 & 10  & Text (100) & Image (1024)&-&-&-&-\\
				Wikipedia & 2866 & 10 & Text (100) & Image (1024) &-&-&-&-\\
				\hline
			\end{tabular}
		\end{table*}
	\end{small}
\end{landscape}

\begin{landscape}
	\begin{small}
		\begin{table*}
			\caption{Mean accuracy and standard deviation of $27$ methods on six multi-view data sets over $10$ random splits of $10\%$ training and 90\% testing.} \label{tab:accuracy}
			\centering
			\begin{tabular}{@{}lccccccr@{}}
				\hline
				Method & Mfeat  & Ads & Caltech101-7 & Caltech101-20 & Scene15& NUS-wide-object & rank\\
				\hline\hline
				MCCA &79.11 $\pm$ 1.36 (21) & 91.05 $\pm$ 1.51 (26) & 88.53 $\pm$ 2.03 (25) & 63.54 $\pm$ 2.58 (25) & 43.41 $\pm$ 1.93 (27) & 34.53 $\pm$ 0.79 (24) & 24.7\\
				MvOPLS &74.29 $\pm$ 1.94 (23) & 92.04 $\pm$ 1.42 (23) & 93.80 $\pm$ 1.04 (20) & 84.75 $\pm$ 1.21 (21) & 55.87 $\pm$ 1.26 (23) & 34.97 $\pm$ 0.37 (20) & 21.7\\
				MvDA &74.05 $\pm$ 2.23 (25) & 91.62 $\pm$ 1.65 (25) & 92.77 $\pm$ 3.63 (23) & 84.09 $\pm$ 1.29 (23) & 55.78 $\pm$ 1.52 (24) & 34.96 $\pm$ 0.37 (22) & 23.7\\
				MvDA-VC &89.06 $\pm$ 1.74 (18) & 93.04 $\pm$ 2.35 (19) & 94.21 $\pm$ 0.82 (19) & 88.76 $\pm$ 1.04 (18) & 75.75 $\pm$ 2.18 (19) & 34.81 $\pm$ 0.52 (23) & 19.3\\
				MvLDA &95.46 $\pm$ 0.80 ( 8) & 92.90 $\pm$ 1.62 (20) & 93.13 $\pm$ 1.15 (21) & 90.53 $\pm$ 0.69 (10) & 92.24 $\pm$ 0.60 (12) & 27.22 $\pm$ 0.70 (27) & 16.3\\
				MvMDA &41.13 $\pm$ 9.31 (27) & 91.67 $\pm$ 1.66 (24) & 77.47 $\pm$ 3.51 (26) & 52.60 $\pm$ 1.40 (26) & 77.15 $\pm$ 1.19 (18) & 31.28 $\pm$ 0.37 (26) & 24.5\\
				MLDA &78.96 $\pm$ 1.32 (22) & 90.92 $\pm$ 1.52 (27) & 88.56 $\pm$ 2.12 (24) & 63.81 $\pm$ 2.55 (24) & 44.96 $\pm$ 1.62 (26) & 34.48 $\pm$ 0.78 (25) & 24.7\\
				GMA &41.14 $\pm$ 9.30 (26) & 93.07 $\pm$ 1.81 (17) & 76.21 $\pm$ 3.63 (27) & 52.18 $\pm$ 1.38 (27) & 77.16 $\pm$ 1.19 (17) & 35.66 $\pm$ 0.57 (18) & 22.0\\
				MvDA-CCA &74.06 $\pm$ 2.23 (24) & 93.28 $\pm$ 1.74 (16) & 92.92 $\pm$ 2.95 (22) & 84.35 $\pm$ 1.28 (22) & 56.40 $\pm$ 1.19 (22) & 34.97 $\pm$ 0.37 (21) & 21.2\\ \hline\hline
				MCCAp &88.41 $\pm$ 1.70 (19) & 92.53 $\pm$ 1.01 (21) & 95.25 $\pm$ 0.45 (16) & 88.82 $\pm$ 0.77 (16) & 72.24 $\pm$ 1.08 (20) & 37.54 $\pm$ 0.41 (16) & 18.0\\
				MvOPLSp &95.23 $\pm$ 0.86 (11) & 95.21 $\pm$ 0.76 (13) & 95.87 $\pm$ 0.32 (14) & 91.54 $\pm$ 0.61 ( 9) & 92.92 $\pm$ 0.52 ( 7) & 37.63 $\pm$ 0.54 (12) & 11.0\\
				MvDAp &95.23 $\pm$ 0.86 (12) & 95.29 $\pm$ 0.71 (12) & 95.99 $\pm$ 0.49 (12) & 91.58 $\pm$ 0.67 ( 7) & 92.95 $\pm$ 0.49 ( 6) & 37.62 $\pm$ 0.53 (14) & 10.5\\
				MvDA-VCp &95.21 $\pm$ 0.88 (14) & 95.32 $\pm$ 0.77 (11) & 95.95 $\pm$ 0.48 (13) & 91.55 $\pm$ 0.65 ( 8) & 92.84 $\pm$ 0.54 ( 8) & 37.63 $\pm$ 0.55 (13) & 11.2\\
				MvLDAp &93.53 $\pm$ 1.17 (17) & 93.05 $\pm$ 0.94 (18) & 95.17 $\pm$ 0.97 (18) & 90.01 $\pm$ 0.76 (12) & 45.83 $\pm$ 5.16 (25) & 35.53 $\pm$ 0.67 (19) & 18.2\\
				MvMDAp &95.32 $\pm$ 0.69 (10) & 95.34 $\pm$ 0.61 (10) & 96.31 $\pm$ 1.04 (10) & 87.41 $\pm$ 1.29 (19) & 92.54 $\pm$ 0.54 (10) & 36.28 $\pm$ 0.31 (17) & 12.7\\
				MLDAp &88.41 $\pm$ 1.69 (20) & 92.49 $\pm$ 1.02 (22) & 95.25 $\pm$ 0.45 (17) & 88.82 $\pm$ 0.78 (17) & 72.24 $\pm$ 1.09 (21) & 37.55 $\pm$ 0.41 (15) & 18.7\\
				GMAp &95.44 $\pm$ 0.75 ( 9) & 95.44 $\pm$ 0.59 ( 8) & 96.73 $\pm$ 0.77 ( 7) & 87.02 $\pm$ 1.20 (20) & 92.38 $\pm$ 0.57 (11) & 38.60 $\pm$ 0.38 (10) & 10.8\\
				MvDA-CCAp &95.22 $\pm$ 0.87 (13) & 95.42 $\pm$ 0.73 ( 9) & 95.86 $\pm$ 0.36 (15) & 91.60 $\pm$ 0.72 ( 6) & 92.78 $\pm$ 0.49 ( 9) & 37.65 $\pm$ 0.52 (11) & 10.5\\\hline\hline
				DMCCA &95.17 $\pm$ 0.64 (15) & 93.95 $\pm$ 0.49 (15) & 96.61 $\pm$ 0.33 ( 8) & 90.28 $\pm$ 0.66 (11) & 81.68 $\pm$ 1.31 (15) & 40.11 $\pm$ 0.35 ( 7) & 11.8\\
				DMvOPLS &95.85 $\pm$ 0.42 ( 2) & 95.76 $\pm$ 0.42 ( 6) & 96.83 $\pm$ 0.37 ( 6) & 92.58 $\pm$ 0.35 ( 2) & 93.31 $\pm$ 0.56 ( 4) & 41.12 $\pm$ 0.29 ( 1) & 3.5\\
				DMvDA &95.61 $\pm$ 0.50 ( 6) & 95.78 $\pm$ 0.47 ( 4) & 96.84 $\pm$ 0.39 ( 4) & 92.57 $\pm$ 0.42 ( 3) & 93.36 $\pm$ 0.50 ( 2) & 41.01 $\pm$ 0.38 ( 3) & 3.7\\
				DMvDA-VC &95.61 $\pm$ 0.50 ( 7) & 95.78 $\pm$ 0.47 ( 5) & 96.84 $\pm$ 0.39 ( 5) & 92.57 $\pm$ 0.42 ( 4) & 93.36 $\pm$ 0.50 ( 3) & 41.01 $\pm$ 0.38 ( 4) & 4.7\\
				DMvLDA &96.35 $\pm$ 0.73 ( 1) & 95.82 $\pm$ 0.36 ( 3) & 97.60 $\pm$ 0.56 ( 1) & 92.72 $\pm$ 0.63 ( 1) & 94.25 $\pm$ 0.24 ( 1) & 40.81 $\pm$ 0.36 ( 5) & 2.0\\
				DMvMDA &95.76 $\pm$ 0.59 ( 3) & 95.91 $\pm$ 0.35 ( 1) & 97.25 $\pm$ 0.54 ( 2) & 89.71 $\pm$ 1.12 (14) & 92.24 $\pm$ 0.35 (13) & 39.27 $\pm$ 0.36 ( 9) & 7.0\\
				DMLDA &95.09 $\pm$ 0.72 (16) & 94.45 $\pm$ 0.34 (14) & 96.45 $\pm$ 0.33 ( 9) & 89.89 $\pm$ 0.73 (13) & 80.64 $\pm$ 0.81 (16) & 39.83 $\pm$ 0.42 ( 8) & 12.7\\
				DGMA &95.74 $\pm$ 0.78 ( 4) & 95.84 $\pm$ 0.43 ( 2) & 96.18 $\pm$ 0.56 (11) & 88.90 $\pm$ 0.94 (15) & 90.70 $\pm$ 0.45 (14) & 40.55 $\pm$ 0.48 ( 6) & 8.7\\
				DMvDA-CCA &95.62 $\pm$ 0.48 ( 5) & 95.57 $\pm$ 0.39 ( 7) & 96.95 $\pm$ 0.50 ( 3) & 92.50 $\pm$ 0.42 ( 5) & 93.30 $\pm$ 0.53 ( 5) & 41.05 $\pm$ 0.33 ( 2) & 4.5\\
				\hline
			\end{tabular}
		\end{table*}
		
	\end{small}
\end{landscape}

The first six multi-view data sets are used for multi-view feature extraction evaluated through multi-class classification. Multiple Features (Mfeat)\footnote{https://archive.ics.uci.edu/ml/datasets/Multiple+Features} and Internet Advertisements (Ads) \footnote{https://archive.ics.uci.edu/ml/datasets/internet+advertisements} are downloaded from UCI machine learning repository, where the descriptions of each view can be found in their documentations. Image datasets Caltech101\footnote{http://www.vision.caltech.edu/Image\_Datasets/Caltech101/}\cite{fei2007learning} and
Scene15\footnote{https://figshare.com/articles/15-Scene\_Image\_Dataset/7007177} \cite{lazebnik2006beyond} are created by applying the following descriptors to each image: CENTRIST \cite{wu2008place}, GIST \cite{oliva2001modeling},
LBP \cite{ojala2002multiresolution}, histogram of oriented gradient (HOG), color histogram (CH),
and SIFT-SPM \cite{lazebnik2006beyond}.
Note that we drop  CH from Scene15 due to the gray-level images, and Caltech101 with two data sets consisting of $7$ and $20$ categories are used by following \cite{deng2015multi}. NUS-wide-object is a web image data set consisting of six pre-computed low-level features \footnote{https://lms.comp.nus.edu.sg/wp-content/uploads/2019/research\\/nuswide/NUS-WIDE.html}.

The last three data sets, TVGraz \cite{khan2009tvgraz}, Wikipedia \cite{rasiwasia2010new} and Pascal \cite{rashtchian2010collecting} are employed for cross-modal retrieval, where the image query is used to retrieve text articles and vice-versa.  As pointed out in \cite{pereira2012regularization}, the three data sets demonstrate different properties. Both image and text classifications are low in accuracy for Pascal. On Wikipedia, image classification has low accuracy, but its text classification accuracy is high. TVGraz has good accuracies for both text and images. These data sets are also used in \cite{pereira2012regularization}, where the training/testing data sets are provided: 1558/500 for TVGraz, 2173/693 for Wikipedia, and 700/300 for Pascal.

\subsection{Compared methods}
We compare the performances of all methods in Section \ref{sec:reformulation} and their nonlinear extensions via deep networks proposed in Section \ref{sec:deep} on the multi-view data sets. All methods for learning linear projections will be evaluated on either the original input data sets or the reduced ones obtained by applying PCA to each view so as to reduce the dimension of each view while retaining $95\%$ energy. Since deep networks can automatically learn the low-dimensional representation, they take in the input data sets without any PCA reduction. Specifically, the compared methods consist of seven existing methods including MCCA \cite{via2007learning}, MvLDA \cite{kan2016multi},  MvDA \cite{kan2015multi}, MvDA-VC \cite{kan2015multi}, MvMDA \cite{cao2017generalized}, MLDA \cite{sun2015multiview}, GMA  \cite{sharma2012generalized}, and two new variants including the simplest MvOPLS denoted by MvOPLS, as the combination with the Tikhonov regularization and MvDA+CCA of (\ref{op:mvda-cca}). For the ease of reference, we add suffix ``p'' to the name of each method for the same method applied to the reduced data via PCA, and add prefix ``D'' for its nonlinear extensions via deep networks. For example associated to MvDA, MvDAp stands for MvDA applied to the PCA reduced data, and DMvDA is the nonlinear extension of MvDA via deep networks. To prevent $\widetilde{C}_{\rm diag} + A$ from being singular in the unified from (\ref{op:ropls-ged}), the Tikhonov regularization (\ref{eq:r-tikh}) is applied to all methods. 

\subsection{Experimental settings}
We evaluate the baseline methods on the multi-view data sets in terms of classification. All methods aim to learn a set of linear/nonlinear projections, which transform the data points of each view to points in the common space. The classification is then conducted in the common space. As presented in \cite{foster2008multi}, the concatenation of the projected points from all views as the new representation of the input instance is proper for use by a regression algorithm, and the main finding about CCA is that there is little loss of predictive power by using the reduced data in a lower dimensional space while the regression problem gains a lower sample complexity due to that the reduced multi-view data resides in $\mathbb{R}^{vk \times n}$. It is worth noting that the new representation of multi-view data is consistent with our proposed framework based on least squares. 

The proposed regularized MvOPLS has its built-in classifier, but it is improper for some variants such as MCCA, MLDA and GMA because their least squares losses are independent of the class labels. To make fair comparisons of all baseline methods in terms of classification performance, we seek an independent classifier for performance evaluation. Among them, linear support vector machines (SVMs) and 1-nearest neighbor classifier have been popularly used in the literature \cite{andrew2013deep,wang2015deep,sharma2012generalized,kan2015multi}. We will evaluate baselines in terms of SVMs since it is consistent with the classifier of the proposed framework. Specifically, the data is split into training and testing sets. Each baseline method takes in a training set, and outputs the learned projections and the new representation of the training set. The classifier is trained on the new representation of the training set. In the testing step, the testing set is first transformed to the common space via the given projections, and then the classifier is applied to make predictions of the testing data. We repeat the experiment for each baseline method over $10$ randomly drawn training and testing sets, and the mean accuracy with standard deviation on testing sets is reported.

Regularized MvOPLS and its variants share some common parameters including the regularization parameters for $\mathcal{R}_{\rm tikh}$ and the dimension $k$ of the common space. In addition, methods including MvDA-VC, MLDA, GMA, MvDA-CCA and their nonlinear versions have another regulating parameter $\lambda$ for an additional regularization term. For simplicity, we set $\gamma_s =\gamma= 10^{-4}, \forall s$ in (\ref{eq:r-tikh}), and the second regularization parameter is set to $\lambda=10^{-2}$ in all experiments. 
The dimension $k$ is an important parameter for all subspace learning methods. Following the convention, we will evaluate all baseline methods over a set of $k$s. For Mfeat data, $k \in \{2, 3, 4, 5, 6\}$ is used since only $6$ morphological features are available. For other data sets, $k \in \{2, 3, 5$:$5$:$50\}$ is used. 
The architecture of deep networks used for all nonlinear methods follows \cite{andrew2013deep}, where the widths of the hidden layers are $500$ and $500$, and there are three layers including the output layer. During the training process, we take the full-batch optimization approach, as suggested in \cite{andrew2013deep}. All nonlinear extensions are implemented in Pytorch \cite{NEURIPS2019_9015} for tensor operations and eigenvalue decompositions. The Adam optimizer is used with the learning rate set to $10^{-3}$, and others are set by default. It is worth noting that this work mainly focuses on the generalized framework and its power to recast existing methods as well as inspire new models, and so the best performance achieved by each method via fine-tuning their corresponding parameters is not the main concern of this paper.

Cross-modal retrieval is different from multi-view feature extraction. After the common space is learned on the training data,  the testing data is used to query each other through the common space. We take the retrieval method proposed in \cite{rasiwasia2010new} with $L_2$ metric to evaluate the performance of each method. 
Following \cite{sharma2012generalized}, we set the latent dimension to $20$ for the last three data sets in Table \ref{tab:data} for cross-modal multimedia retrieval. 

\begin{figure}
	\centering
	\begin{tabular}{@{}c@{}c@{}}
		\hline
		Caltech101-20 & Scene15\\
		\hline
		\includegraphics[width=0.5\textwidth]{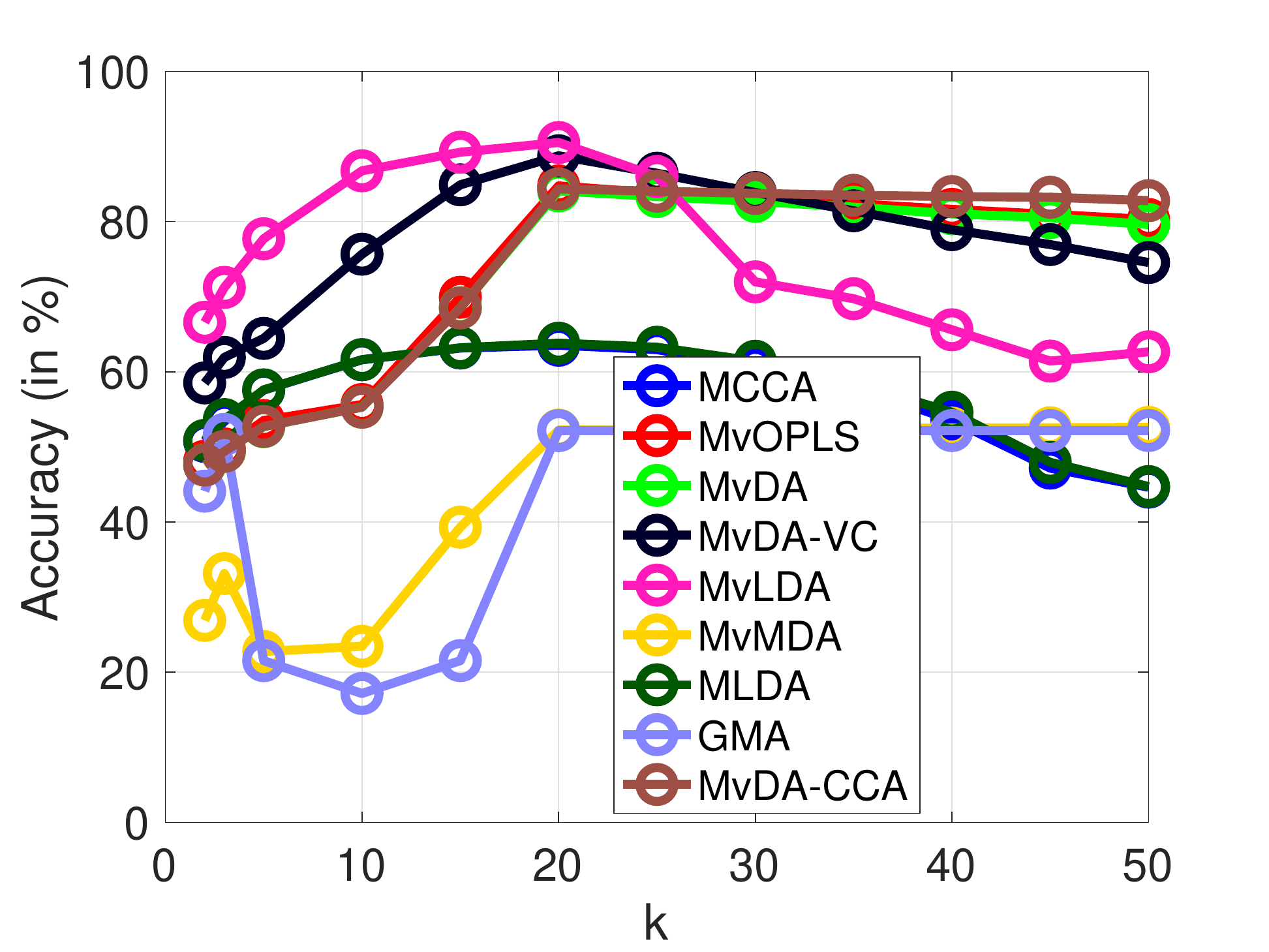} & \includegraphics[width=0.5\textwidth]{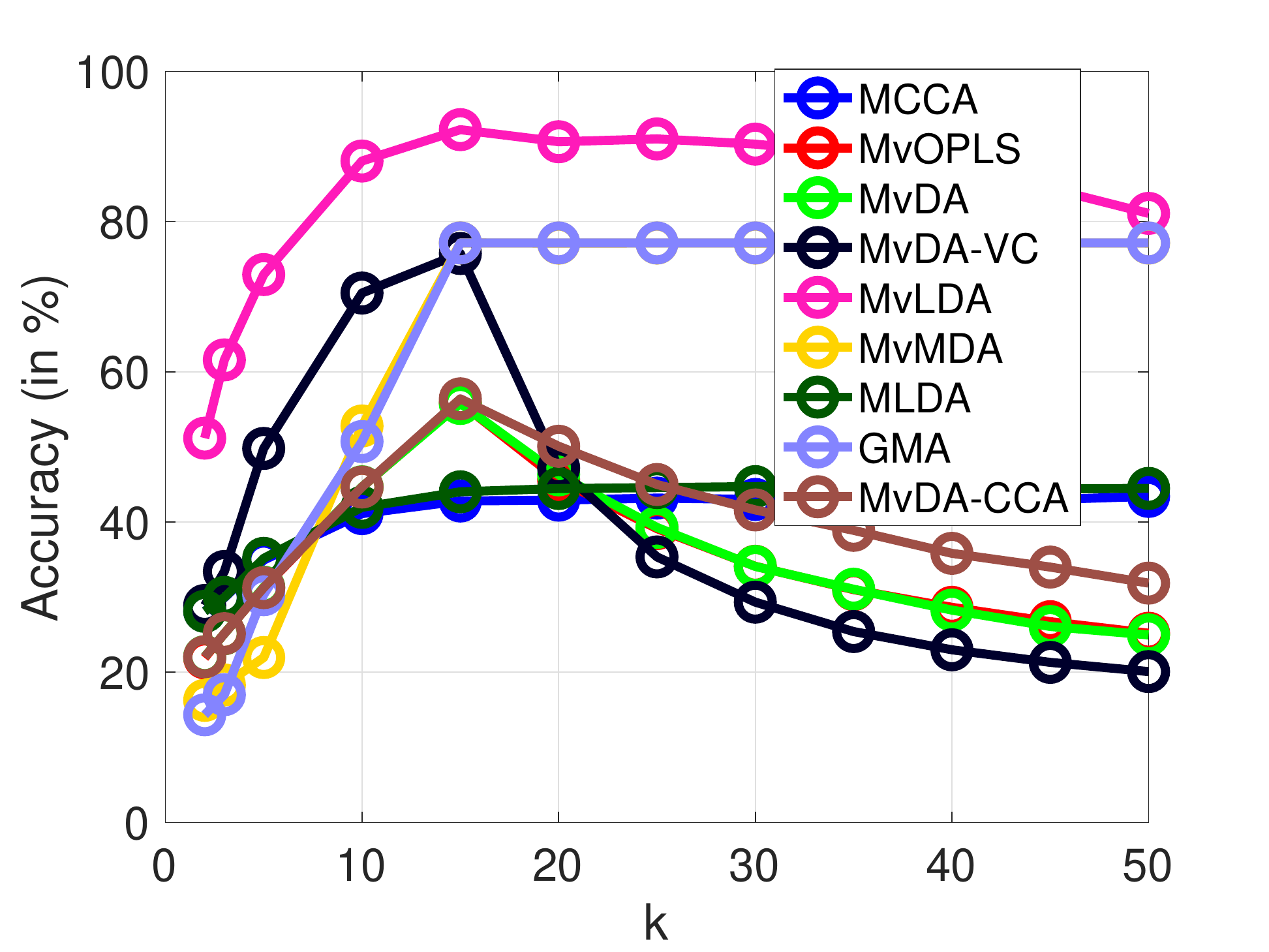}\\
		\includegraphics[width=0.5\textwidth]{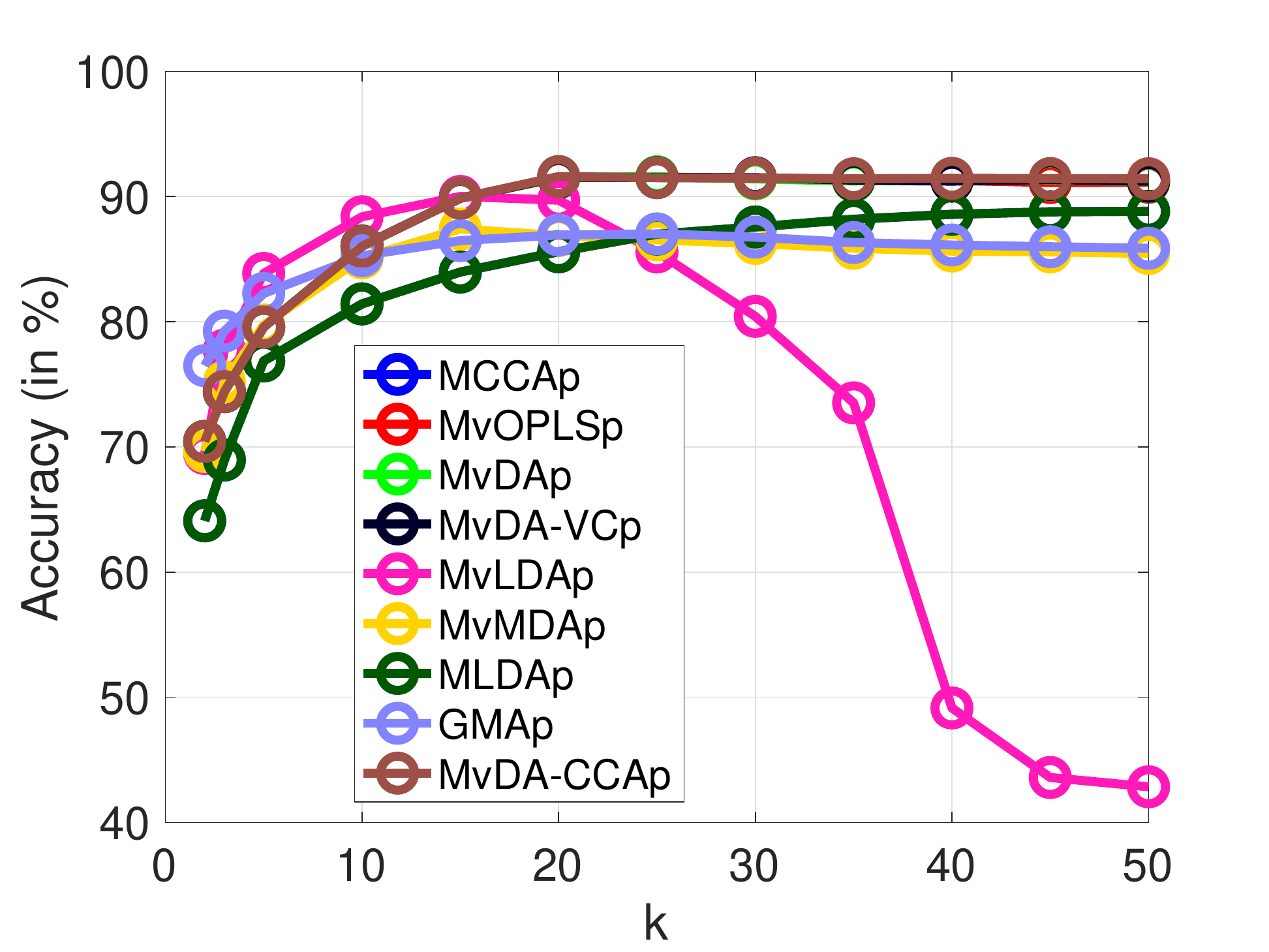} & 
		\includegraphics[width=0.5\textwidth]{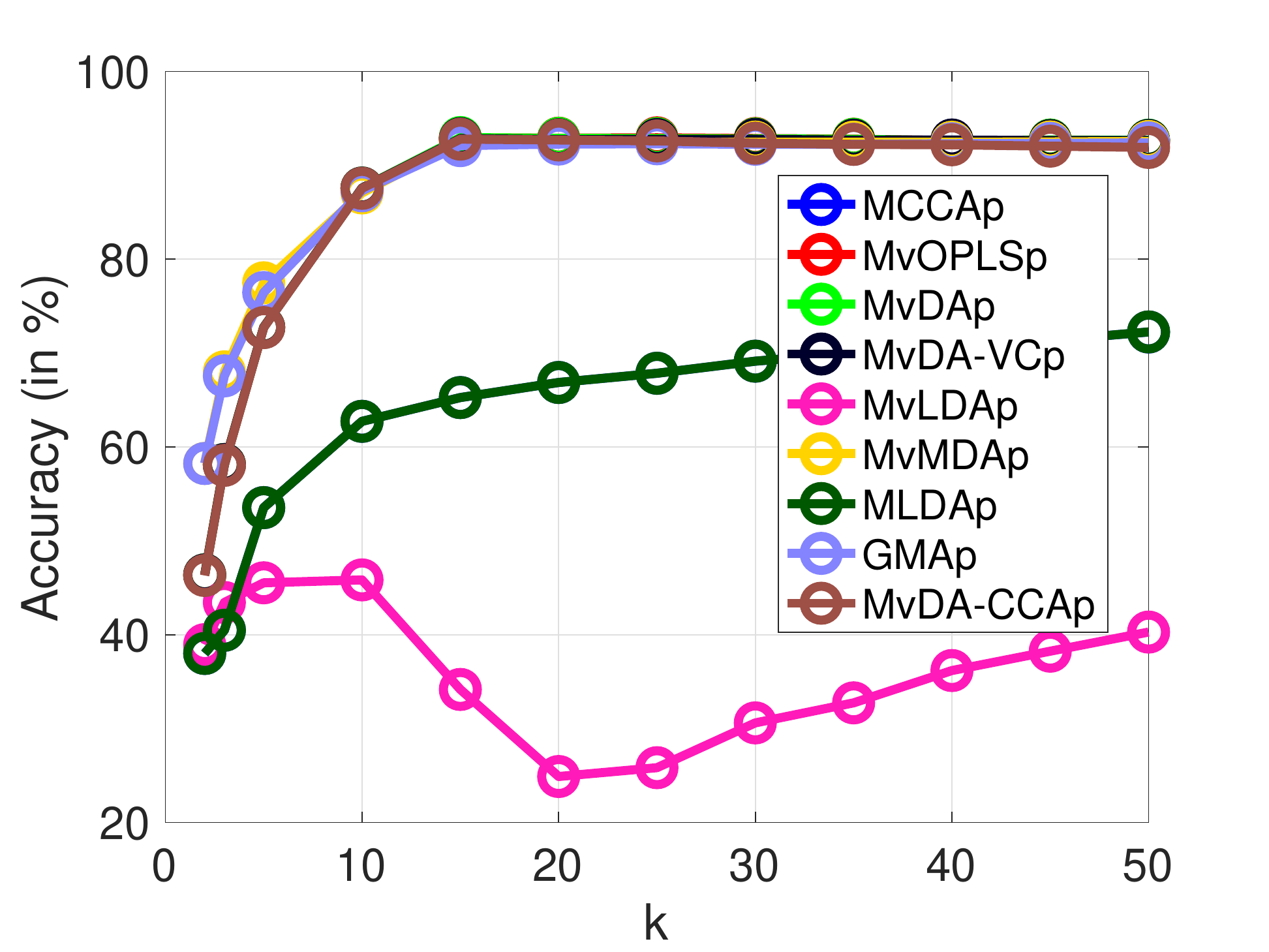}\\
		\includegraphics[width=0.5\textwidth]{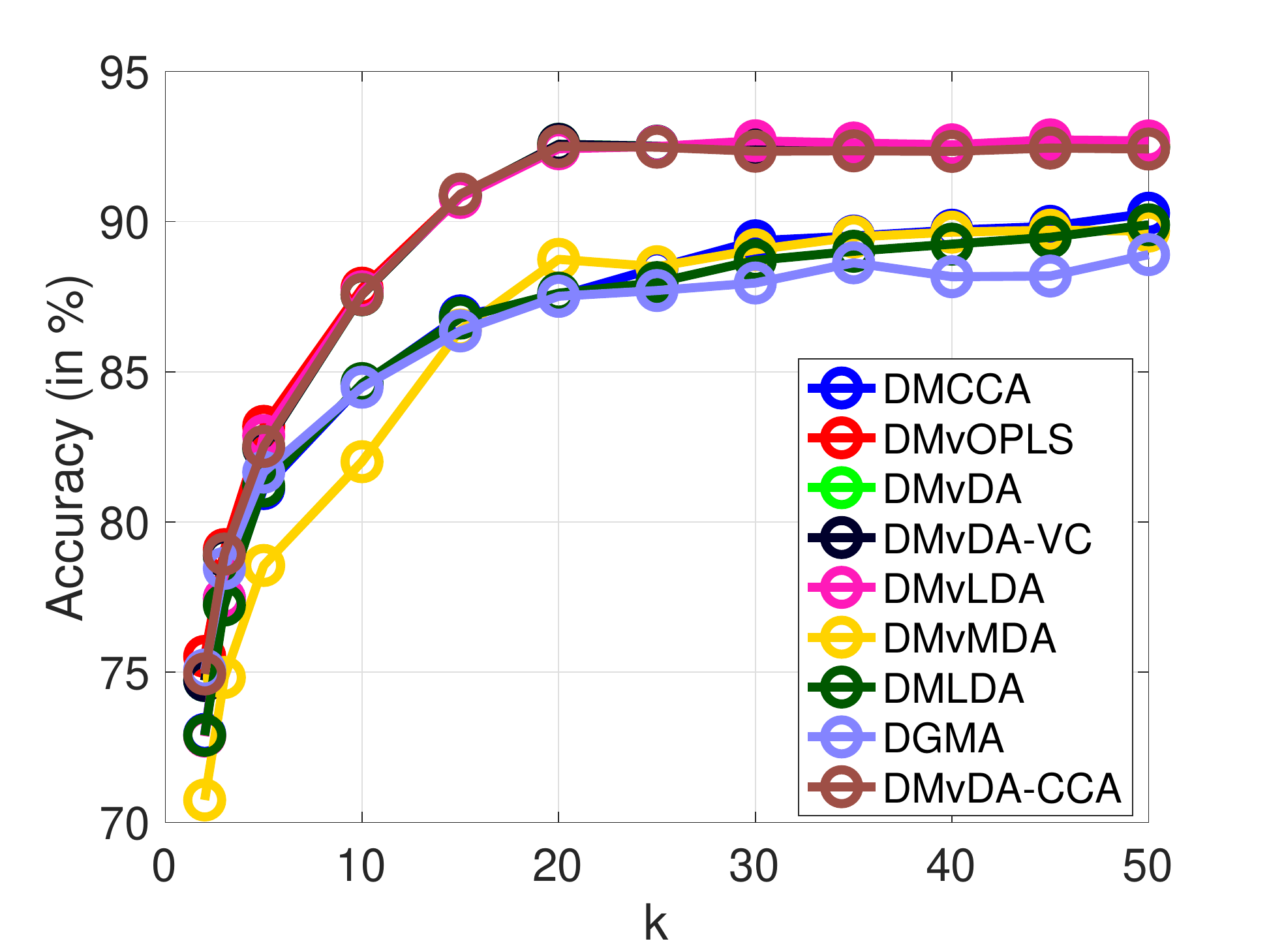} & 
		\includegraphics[width=0.5\textwidth]{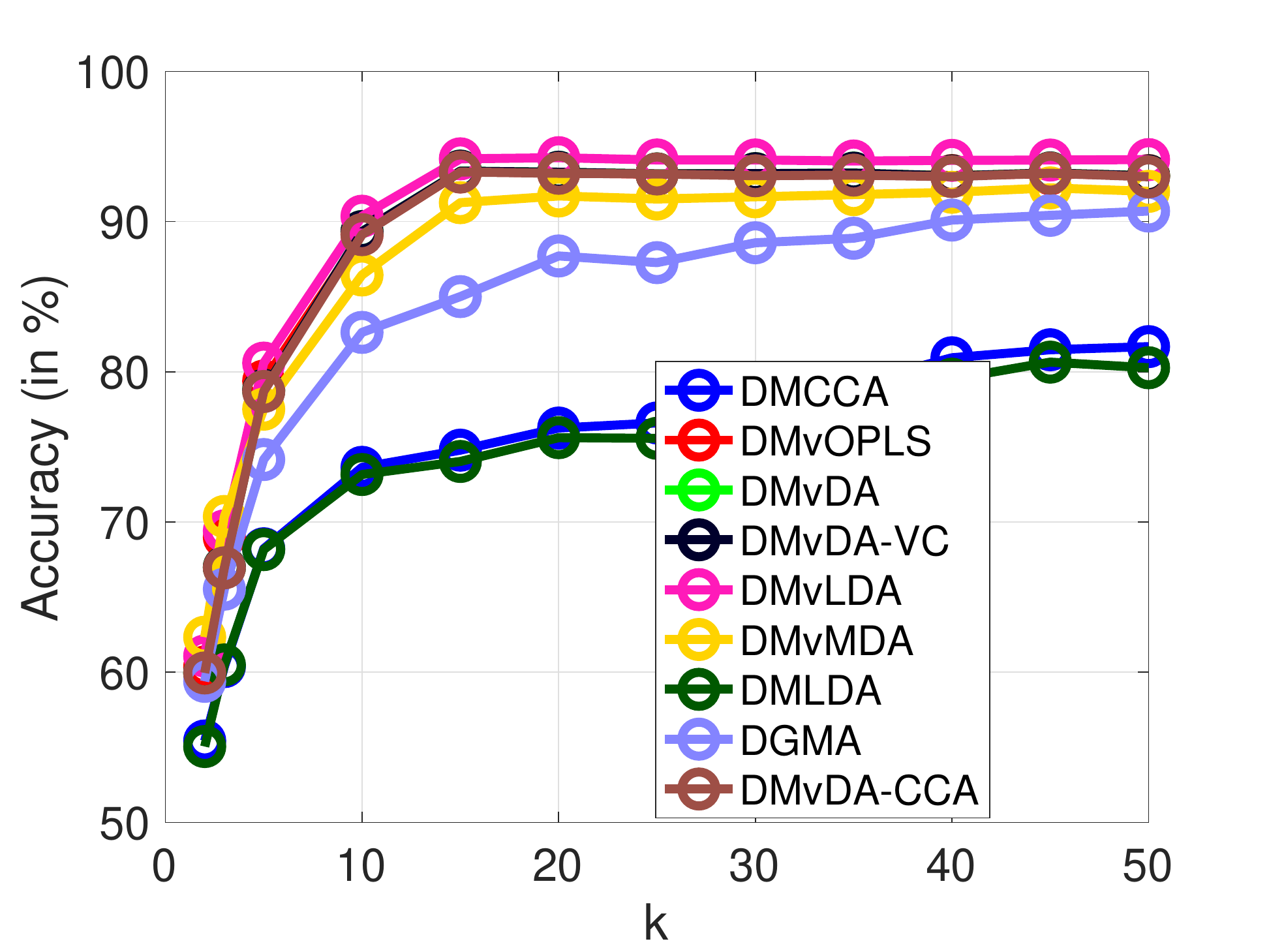}
	\end{tabular} 
	\caption{Accuracies of 27 methods on Caltech101-20 and Scene15 by varying $k$ with $10\%$ training  and $90\%$ testing data split.}\label{fig:k}
\end{figure}

\begin{figure}
	\centering
	\begin{tabular}{@{}c@{}c@{}}
		\hline
		Caltech101-7 & Mfeat\\
		\hline
		\includegraphics[width=0.5\textwidth]{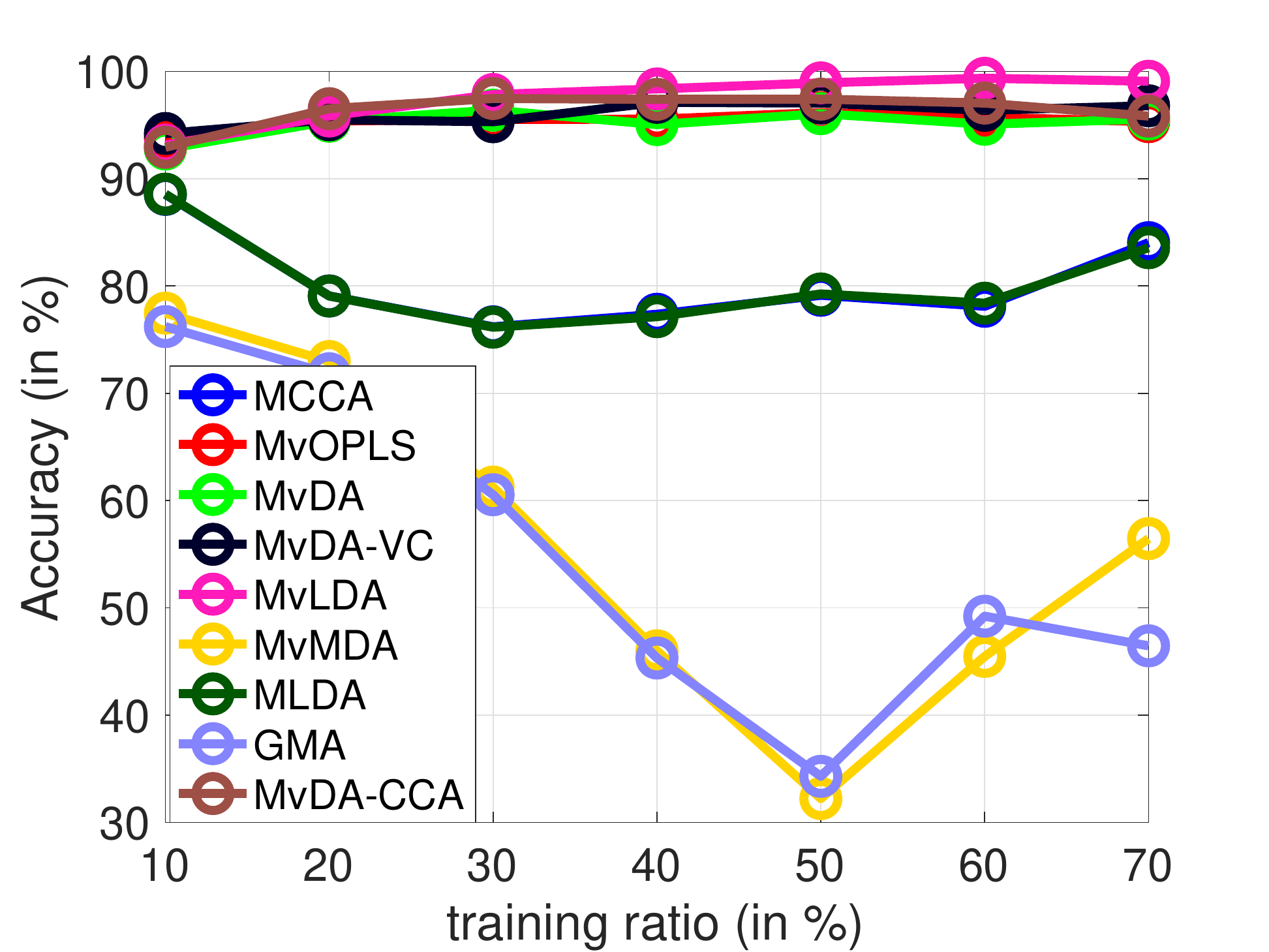} & \includegraphics[width=0.5\textwidth]{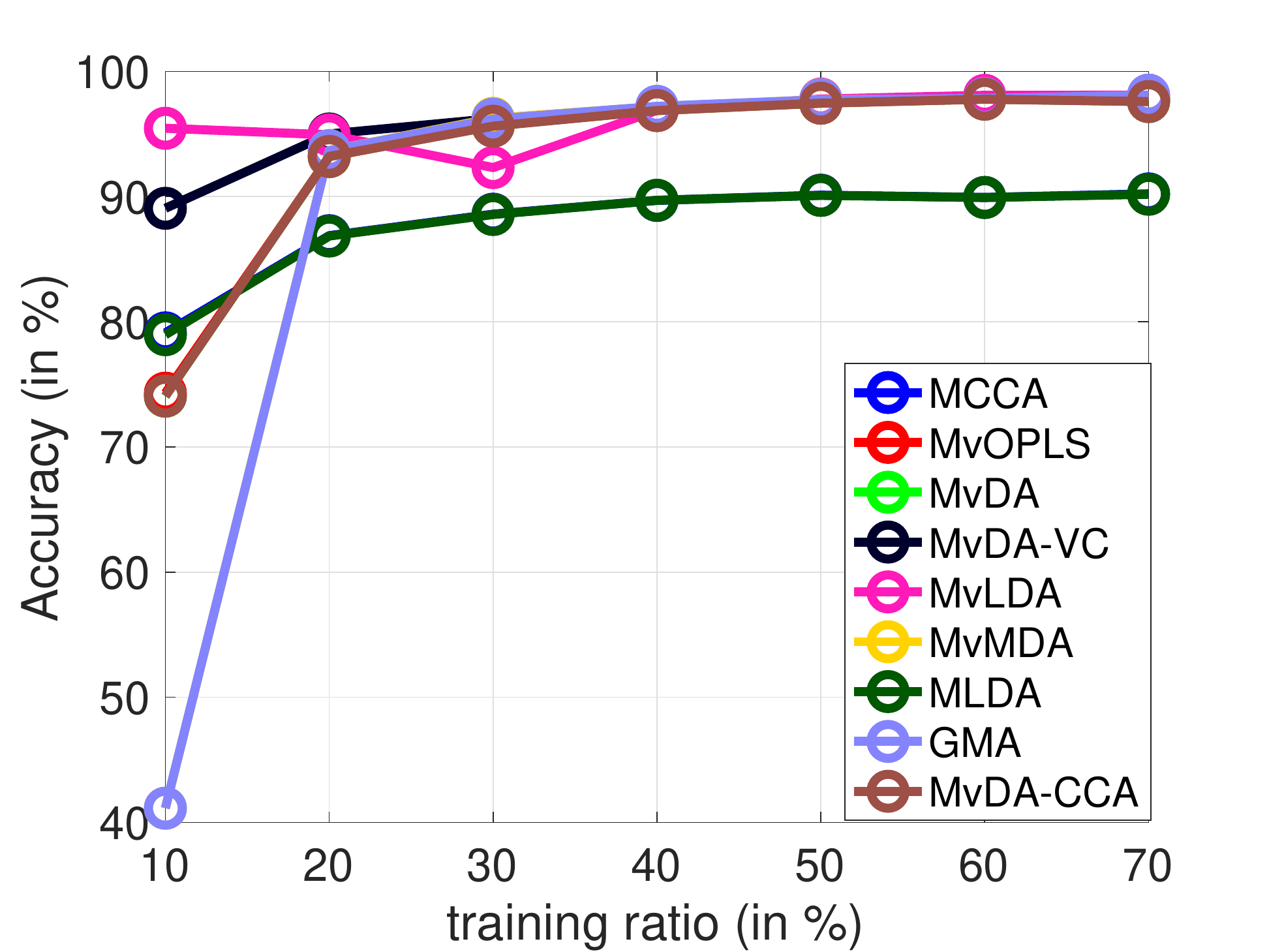}\\
		\includegraphics[width=0.5\textwidth]{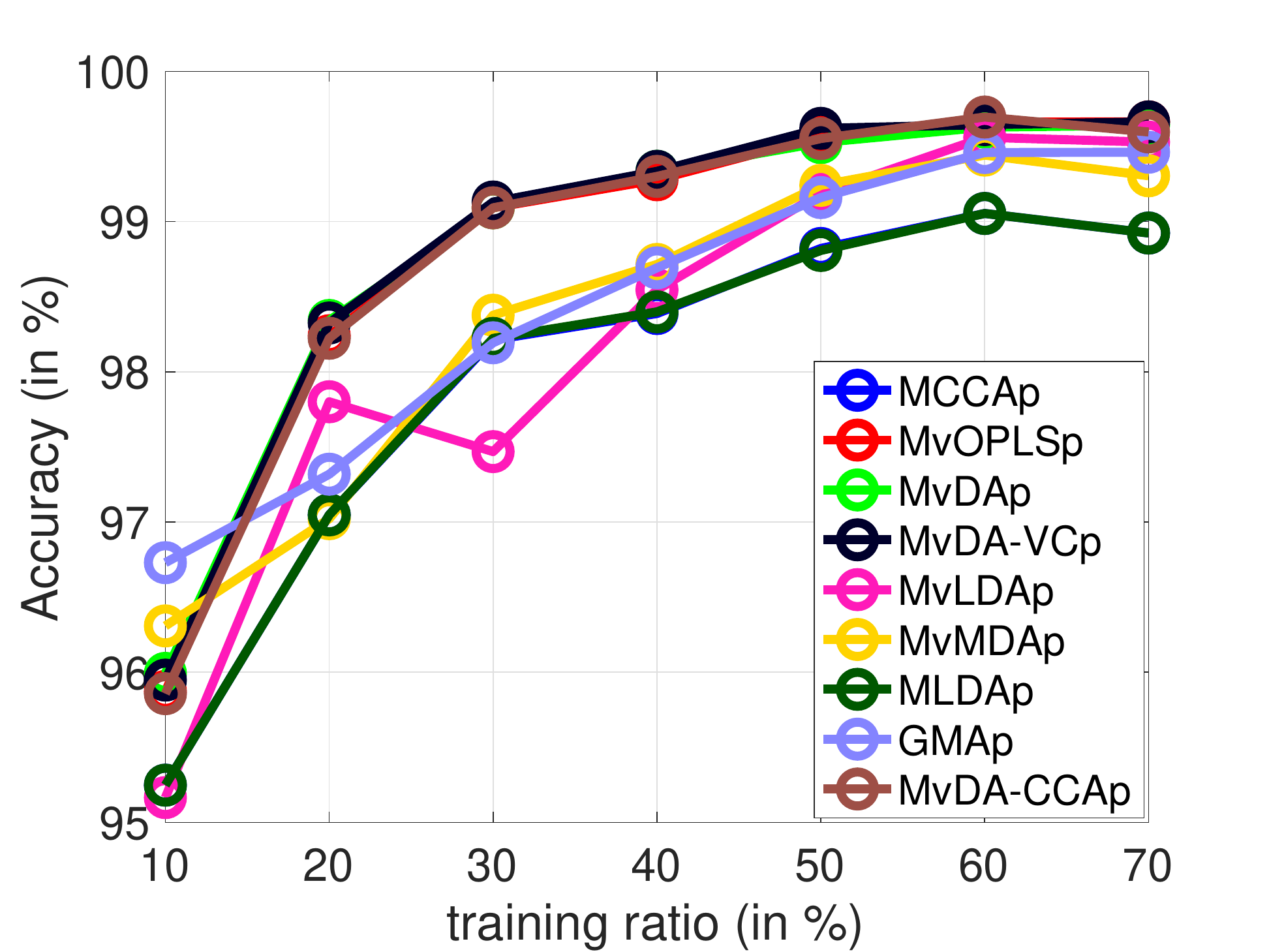} & 
		\includegraphics[width=0.5\textwidth]{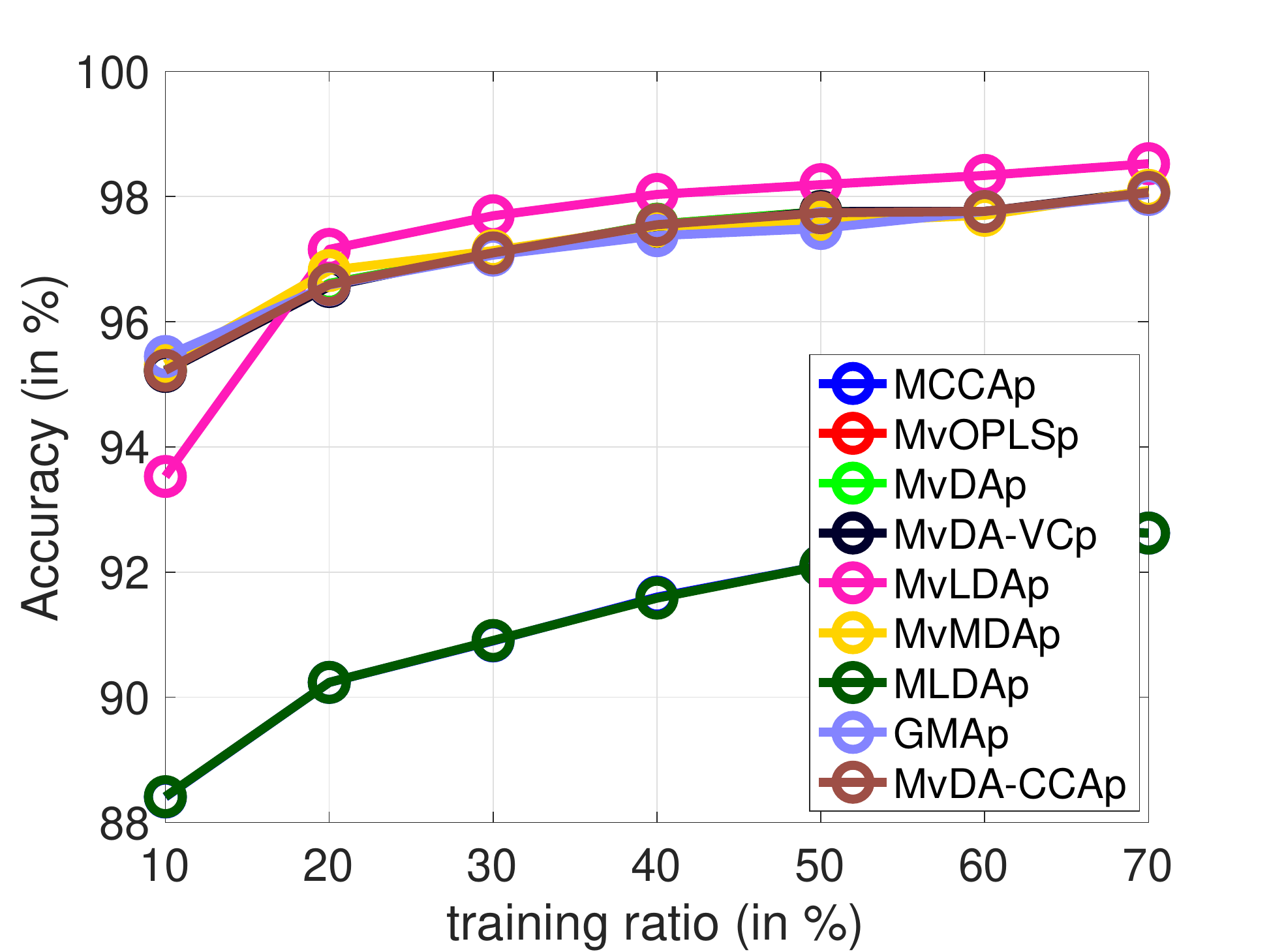}\\
		\includegraphics[width=0.5\textwidth]{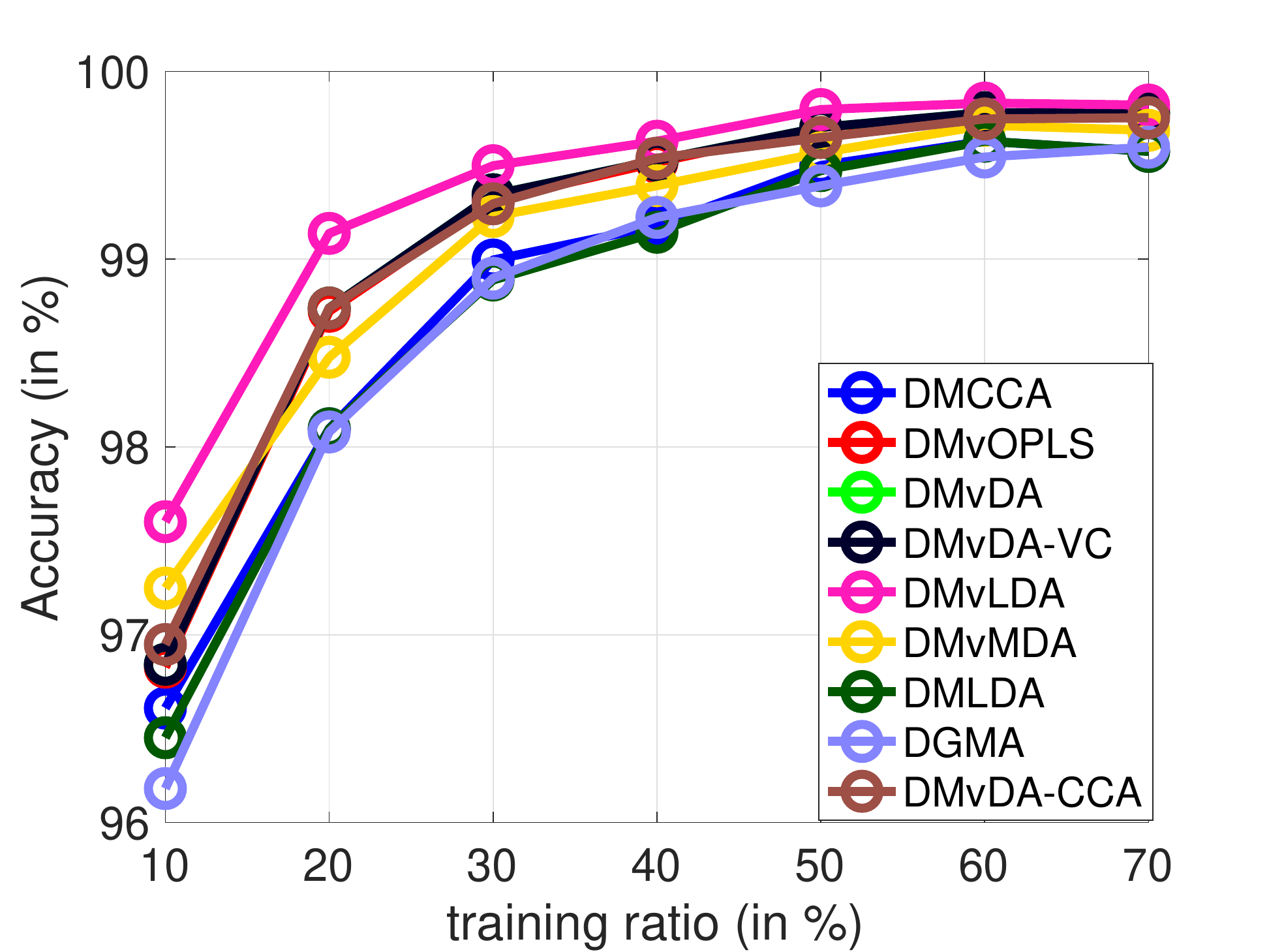} & 
		\includegraphics[width=0.5\textwidth]{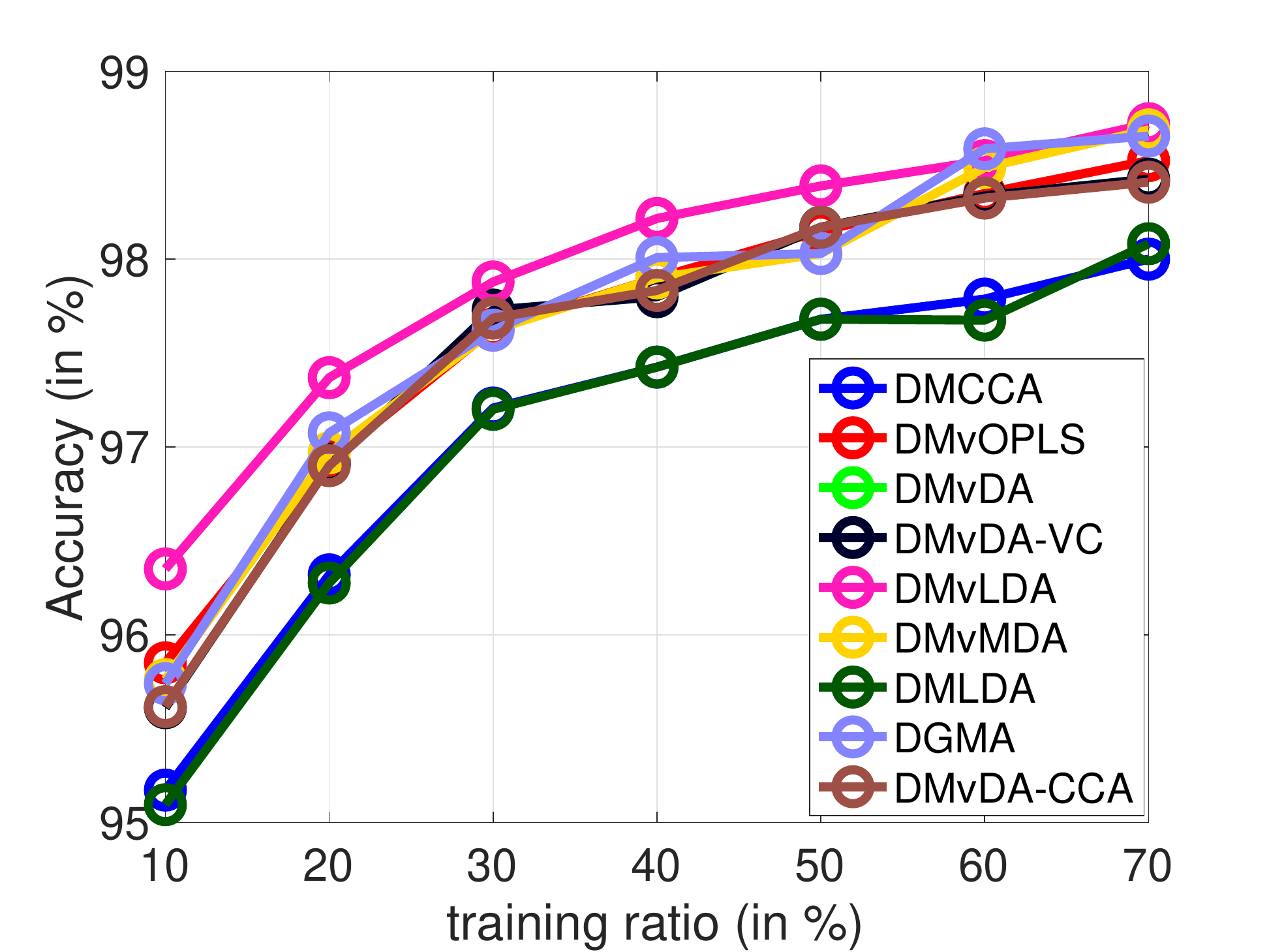}
	\end{tabular} \vspace{-0.15in}
	\caption{Accuracies of 27 methods on Caltech101-7 and Mfeat by varying training data ratio from $10\%$ to $70\%$.}\label{fig:train}
\end{figure}

\subsection{Performance  evaluation via classification} \label{sec:classification}

The classification performances of all methods on the first six data sets in Table \ref{tab:data} are compared from three different perspectives: the overall best accuracy of each method, the accuracy by varying the dimension of the common space, and the best accuracy by varying the training ratios.

\subsubsection{Overall classification performance} \label{sec:overall-class}
We first evaluate nine methods and their three different variants by comparing their best accuracies over all $k$s with $10\%$ training and $90\%$ testing split of data, and the results are shown in Table \ref{tab:accuracy}. We have the following observations: 1) supervised methods significantly outperform unsupervised MCCA; 2) methods (with suffix ``p'') trained on data of reduced dimensions have better accuracies than the same methods on original features  (without suffix ``p'' and prefix ``D''), and nonlinear transformations via deep networks (with suffix ``D'') perform better than the same linear methods (with or without suffix ``p''); 3) among the supervised MvOPLS variants, GMA and MLDA demonstrate relatively worse results than others, and that is possibly caused by the model assumption that their least squares in MvOPLS are built on the unlabeled data. 

To compare all methods over the six data sets, we rank all methods in terms of their accuracies on each data set, from $1$  to $27$ with $1$ being the best  and $27$ being the worst.  
The average ranking over the six data sets is reported to measure the overall performance of each method. The top four methods are DMvLDA, DMvOPLS, DMvDA, and DMvDA-CCA, where the two newly proposed models, DMvOPLS and DMvDA-CCA  are in the  top $2$ and $4$.

\subsubsection{Impact of common subspace}
The impact of common subspace on the performance of classification is evaluated by varying dimension $k$ of the subspace in a given range. Results of the $27$ methods on Caltech101-20 and Scene15 are shown in Fig. \ref{fig:k}. It can be observed that deep variants  consistently achieve better accuracy as $k$ increases. However, those models on original data can behave somehow unpredictable. Among them, MvLDAp behaves very different from other counterparts. In any case, the deep variants produce consistently better results than their two other variants. 

\begin{figure}
	\centering
	\begin{tabular}{@{}c@{}c@{}}
		\hline
		Caltech101-20 & Ads\\
		\hline
		\includegraphics[width=0.5\textwidth]{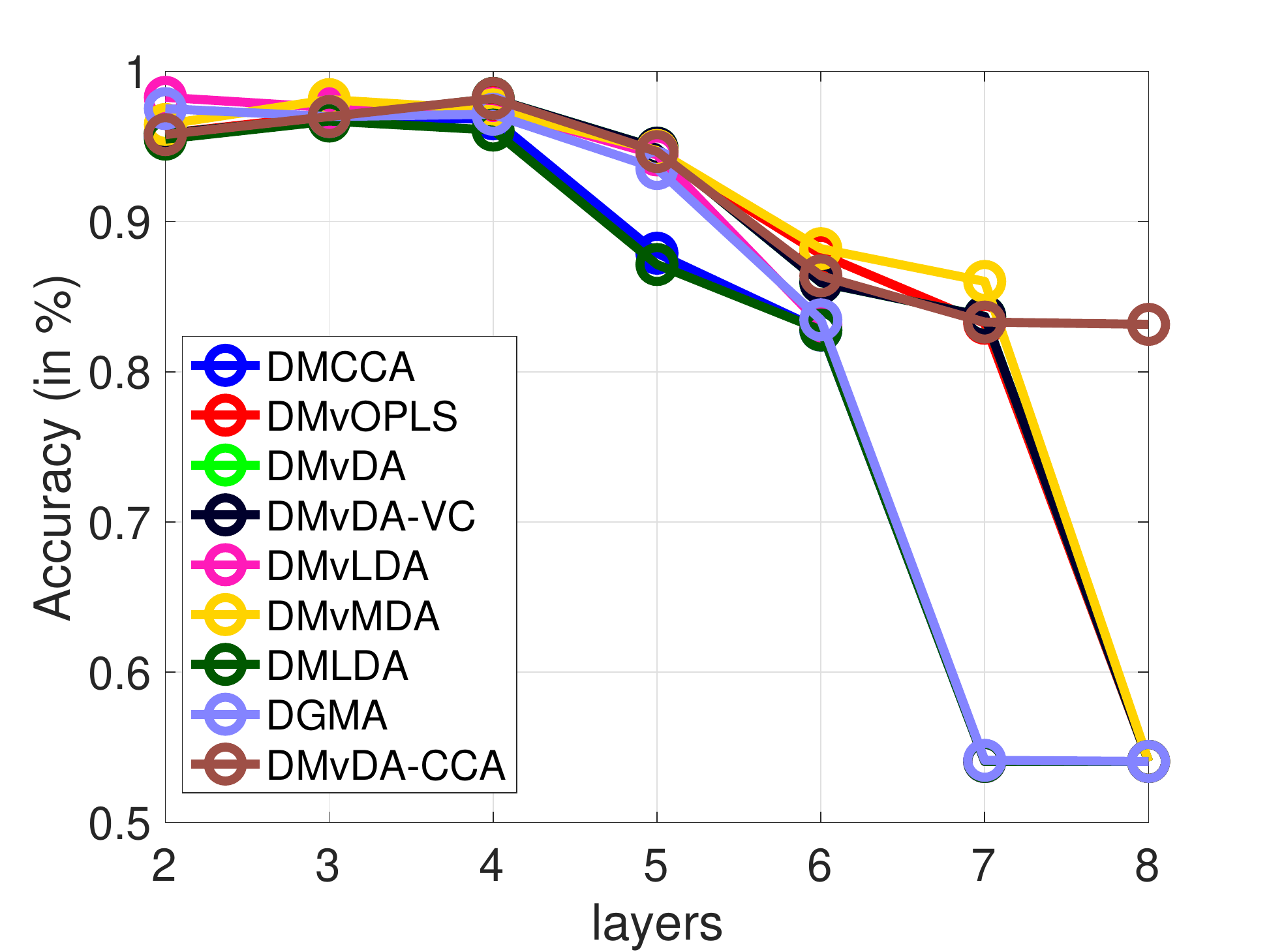} & \includegraphics[width=0.5\textwidth]{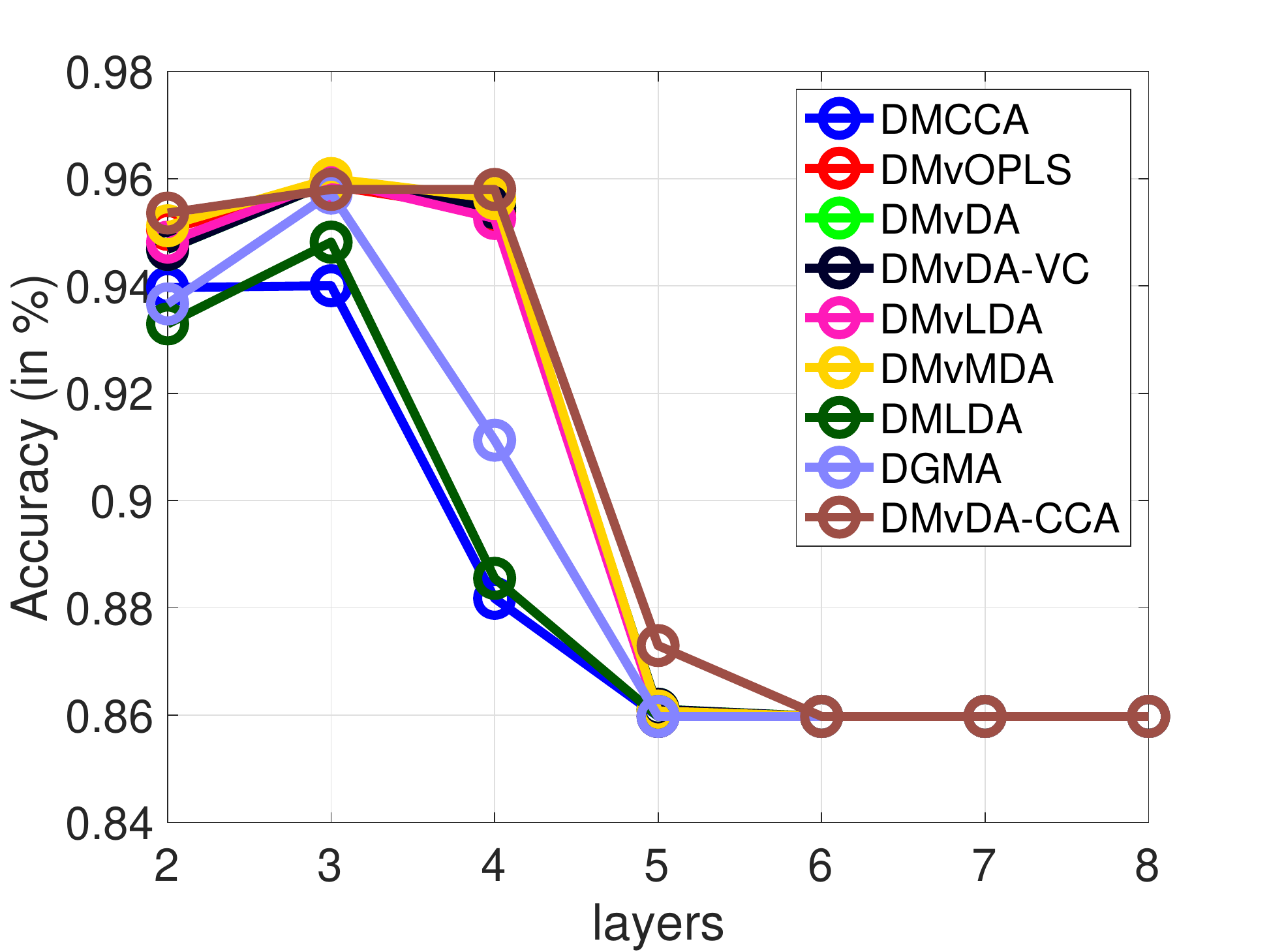}\\
		(a) Sigmoid & (b) Sigmoid\\
		\includegraphics[width=0.5\textwidth]{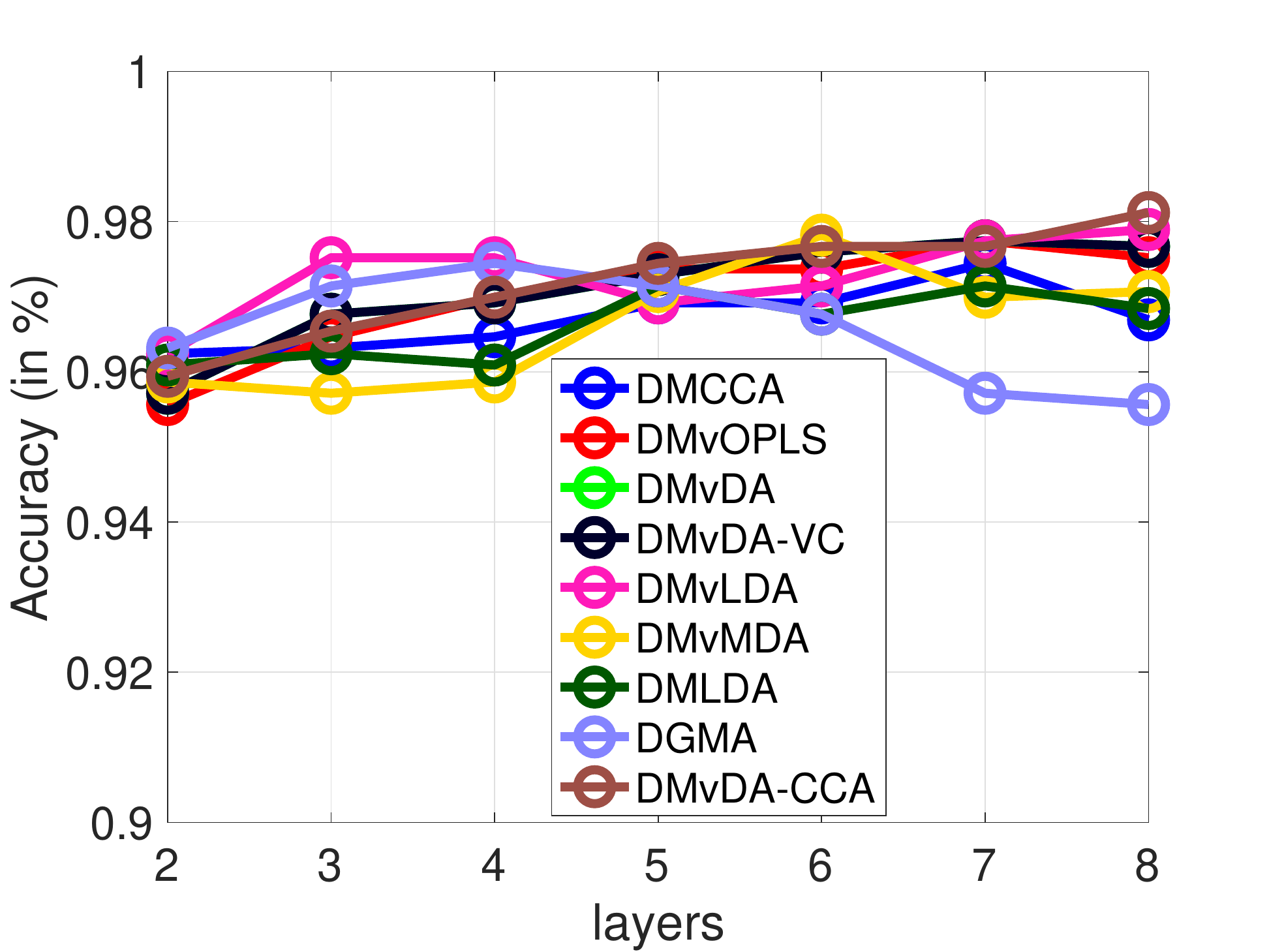} & 
		\includegraphics[width=0.5\textwidth]{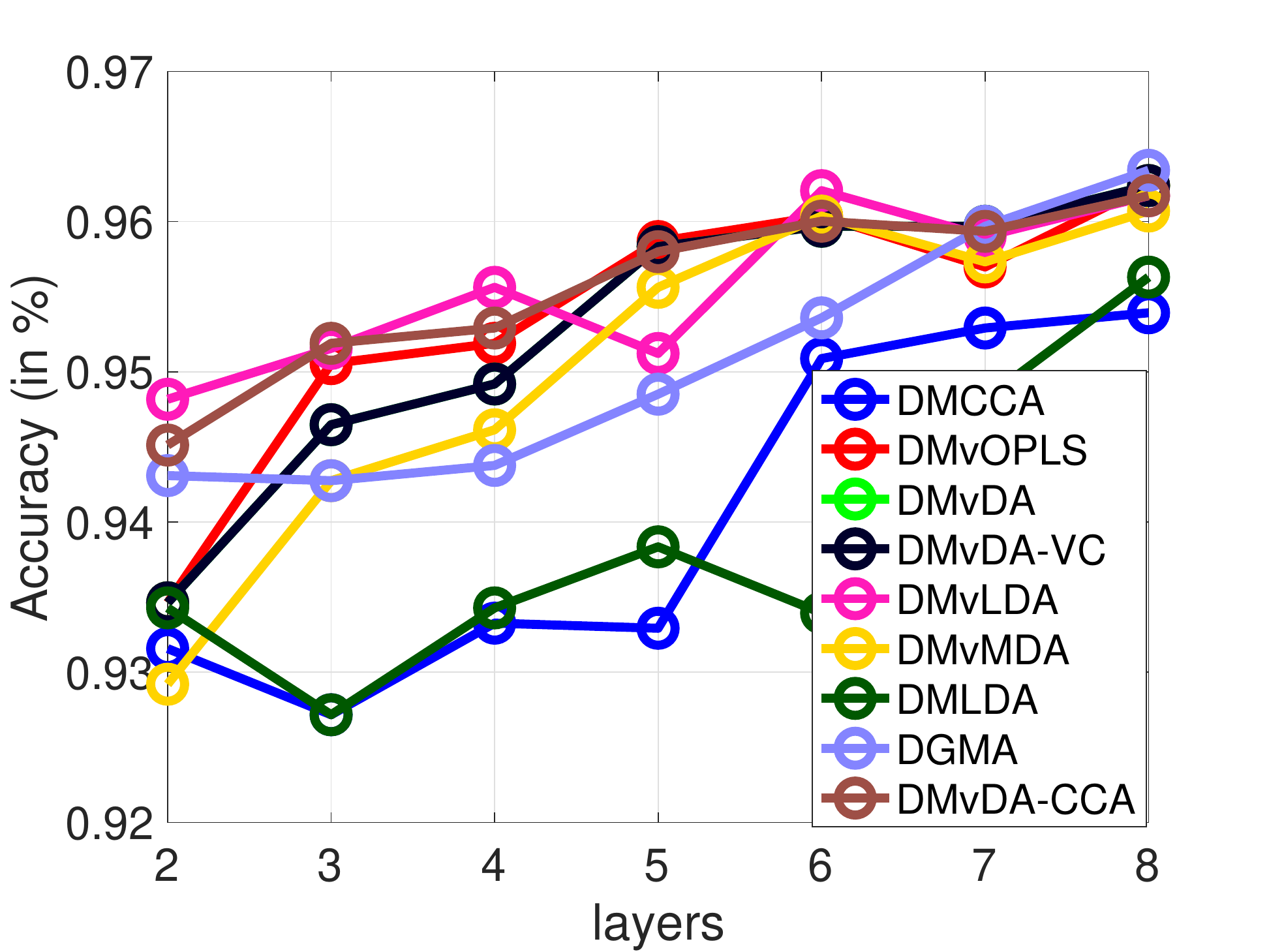}\\
		(c) Tanh & (d) Tanh 
	\end{tabular} \vspace{-0.1in}
	\caption{Accuracies of 9 deep network methods on Caltech101-7 and Ads as the number of layers varies from $2$ to $8$.}\label{fig:layer}
\end{figure} 

\begin{figure}
	\centering
	\begin{tabular}{@{}c@{}c@{}}
		\includegraphics[width=0.5\textwidth]{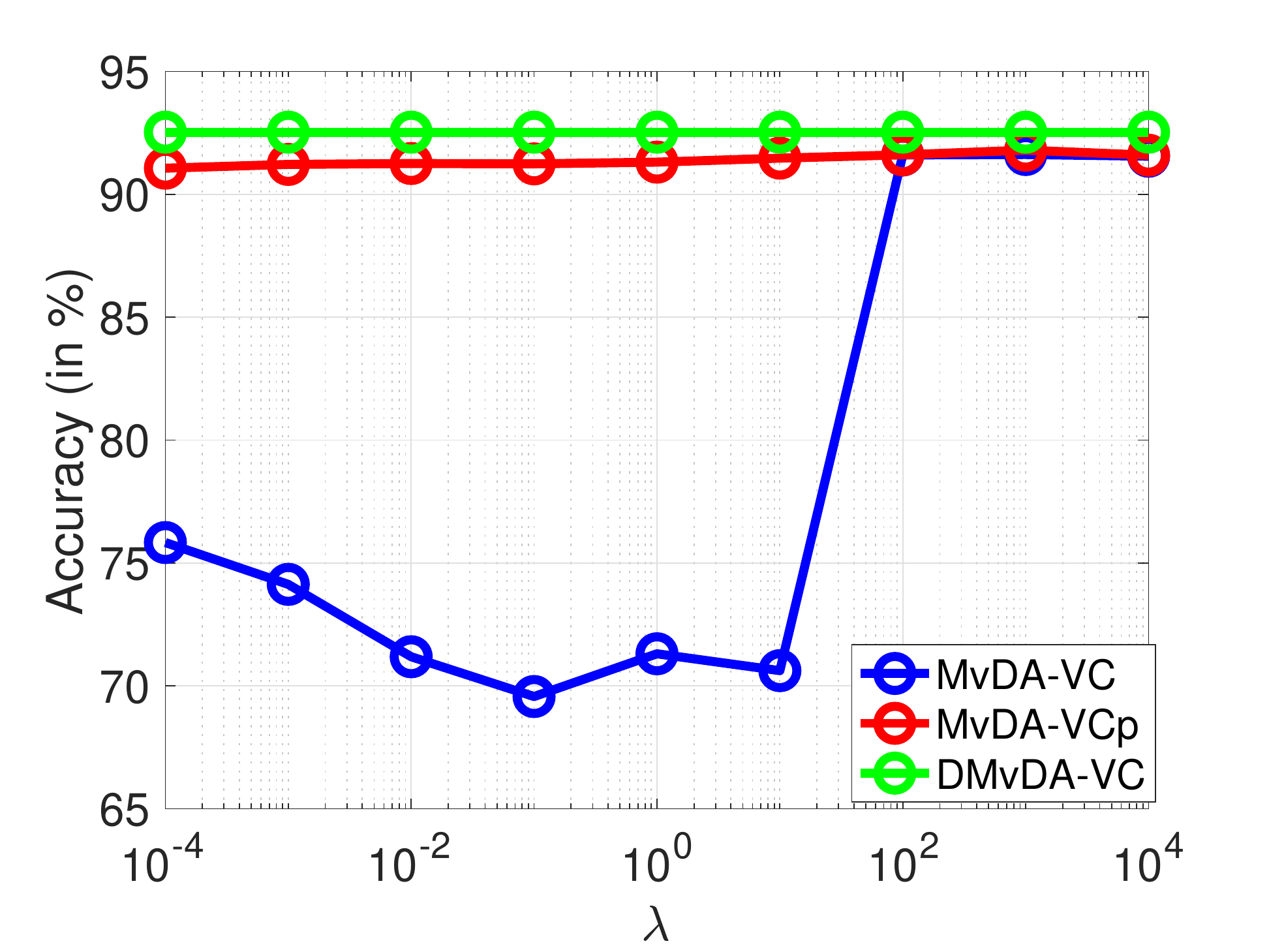} & \includegraphics[width=0.5\textwidth]{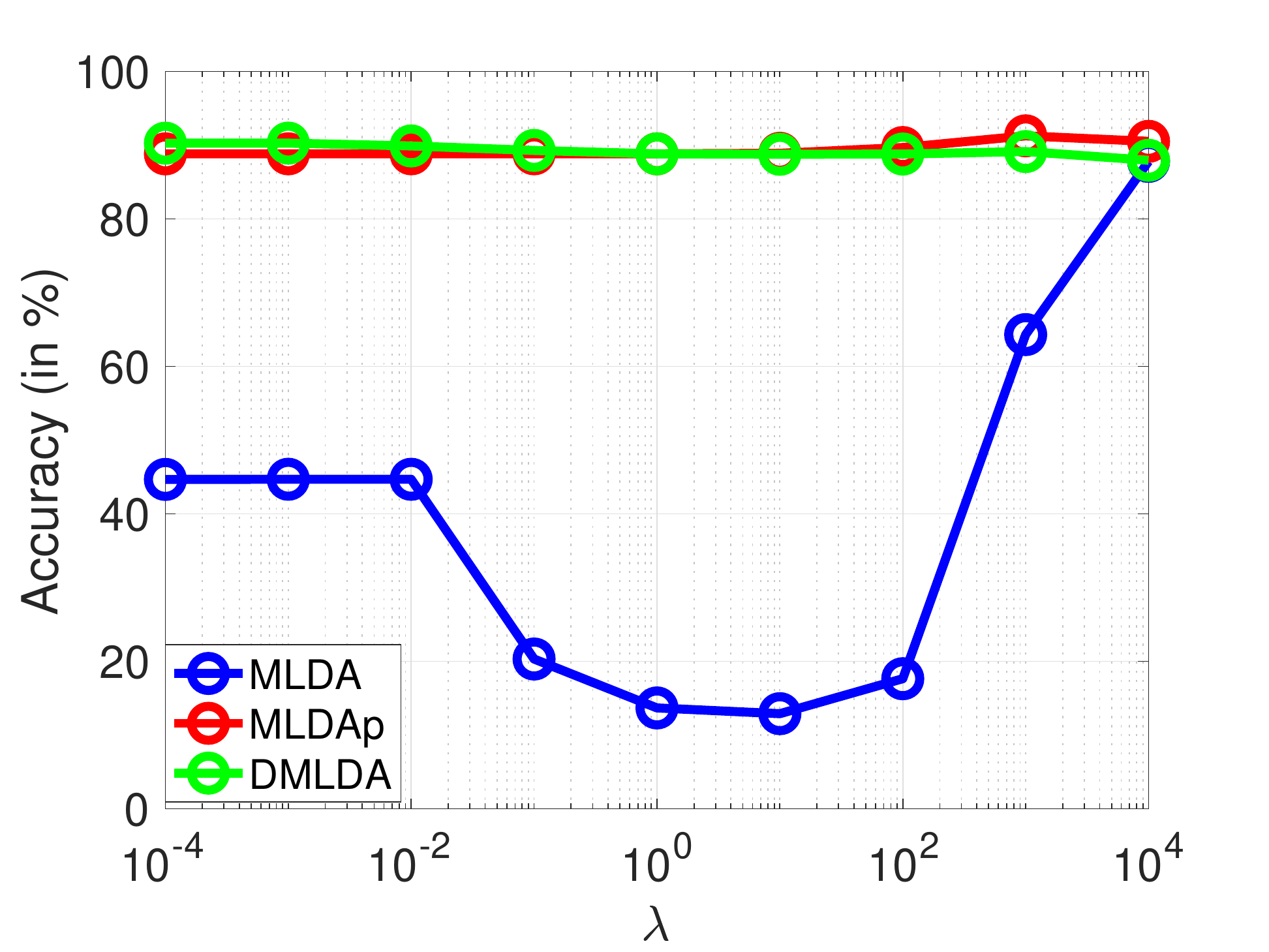}\\
		(a) MvDA-VC & (b) MLDA\\
		\includegraphics[width=0.5\textwidth]{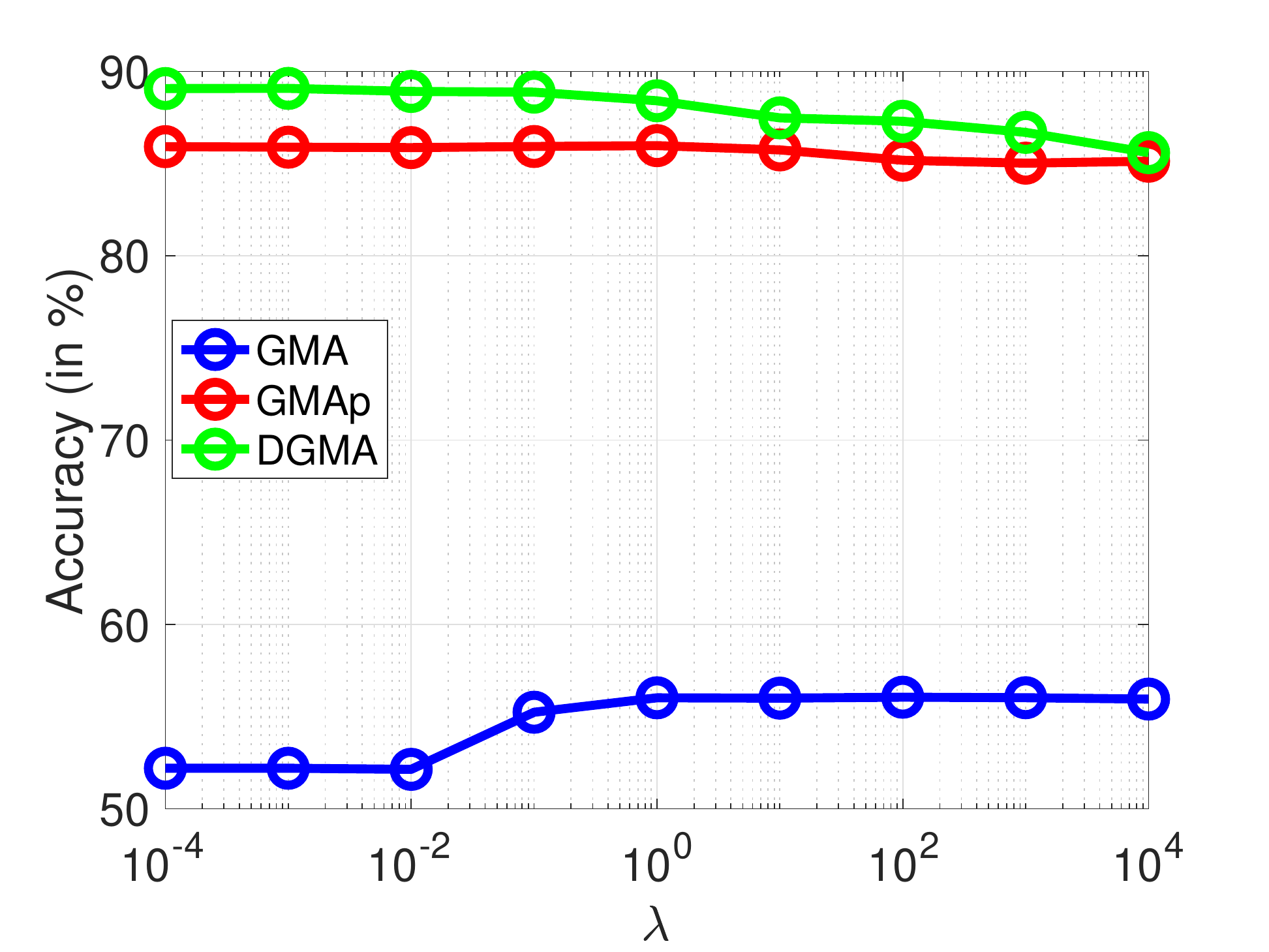} & 
		\includegraphics[width=0.5\textwidth]{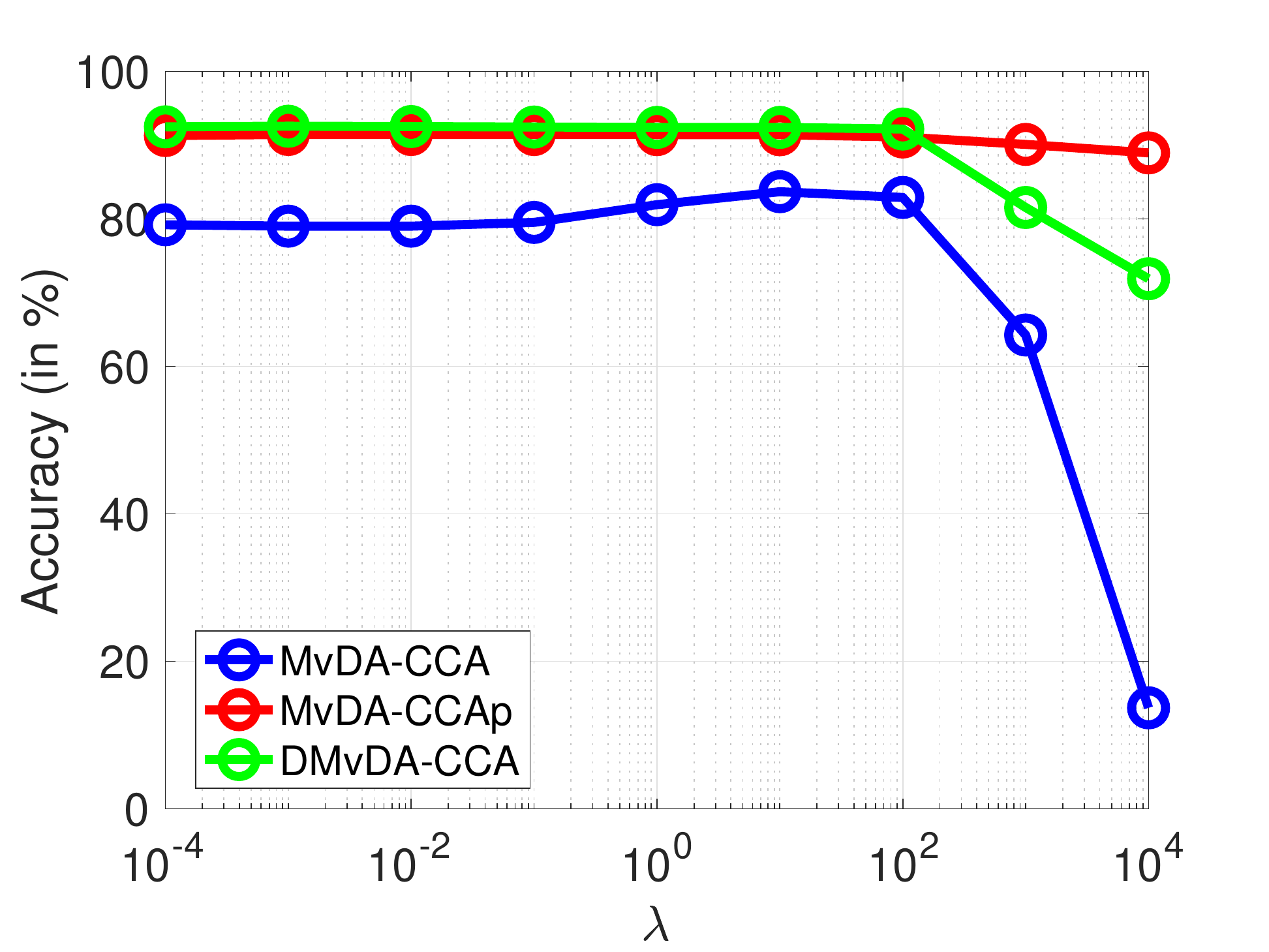}\\
		(c) GMA & (d) MvDA-CCA 
	\end{tabular} \vspace{-0.1in}
	\caption{Parameter sensitivity analysis of four methods on Caltech101-20 by varying parameter $\lambda$ in a wide range.}\label{fig:params}
\end{figure} 

\subsubsection{Impact of the size of training data}
We further evaluate the impact of training ratio on the performance of classification. The results are shown in Fig. \ref{fig:train} where the training ratio is varied from $10\%$ to $70\%$ for the $27$ methods on Mfeat and Caltech101-7. We observe that deep variants perform consistently better as training ratio increases. Methods of other two variants show the similar trends, but with lower accuracy compared to deep variants.

\subsubsection{Impact of the layers of deep networks}
We also evaluate the sensitivity of MvOPLS variants with respect to the depth of networks from $2$ to $8$ layers. Each layer consists of a linear layer with a nonlinear activation function. Here, two functions: Sigmoid and Tanh are used. The widths of all layers are set to $500$. The results on two data sets Caltech101-7 and Ads are shown in Fig. \ref{fig:layer}. With both activation functions, DGMA and DMLDA show a bit worse accuracy, while the proposed DMvDA-CCA performs well in general. For Sigmoid, it is shown that the best performance occurs near 3-4 layers, but the accuracy drops significantly when the number of layers becomes large. This has been observed in literature. However, Tanh generally requires more layers to reach similar performance and continues to improve as the number of layers increases. 

\begin{table}
	\caption{mAP scores of $27$ methods on three data sets, where the best results are highlighted in bold font.} \label{tab:retreival}
	\vspace{-0.1in}
	\begin{tabular}{@{}l@{}|@{}c@{\hspace{0.4em}}c@{\hspace{0.4em}}c@{}@{\hspace{0.4em}}c@{\hspace{0.4em}}c@{\hspace{0.4em}}c@{}@{\hspace{0.4em}}c@{\hspace{0.4em}}c@{\hspace{0.4em}}c@{}} 
		\hline
		\multirow{2}{*}{Method}&\multicolumn{3}{c}{Pascal} & \multicolumn{3}{c}{TVGraz} & \multicolumn{3}{c}{Wikipedia}\\\cline{2-10} 
		& Image & Text & Ave & Image & Text & Ave & Image & Text & Ave \\\hline
		
		MCCA& 0.102& 0.085& 0.093& 0.191& 0.282& 0.237& 0.144& 0.174& 0.159\\
		MvOPLS& 0.132& 0.069& 0.101& 0.201& 0.354& 0.278& 0.142& 0.219& 0.181\\
		MvDA& 0.131& 0.070& 0.101& 0.203& 0.351& 0.277& 0.146& 0.219& 0.183\\
		MvDA-VC& 0.134& 0.069& 0.102& 0.141& 0.348& 0.244& 0.131& 0.196& 0.163\\
		MvLDA& 0.070& 0.068& 0.069& 0.148& 0.225& 0.186& 0.130& 0.138& 0.134\\
		MvMDA& 0.159& 0.067& 0.113& 0.163& 0.359& 0.261& 0.140& 0.224& 0.182\\
		MLDA& 0.101& 0.085& 0.093& 0.231& 0.270& 0.250& 0.163& 0.164& 0.164\\
		GMA& 0.159& 0.067& 0.113& 0.163& 0.346& 0.255& 0.135& 0.183& 0.159\\
		MvDA-CCA& 0.131& 0.068& 0.099& 0.271& 0.451& 0.361& 0.153& 0.235& 0.194\\\hline
		
		MCCAp& 0.115& 0.108& 0.111& 0.191& 0.287& 0.239& 0.144& 0.175& 0.159\\
		MvOPLSp& 0.136& 0.154& 0.145& 0.171& 0.309& 0.240& 0.130& 0.167& 0.149\\
		MvDAp& 0.136& 0.153& 0.145& 0.178& 0.327& 0.253& 0.132& 0.180& 0.156\\
		MvDA-VCp& 0.137& 0.154& 0.145& 0.138& 0.340& 0.239& 0.125& 0.193& 0.159\\
		MvLDAp& 0.073& 0.079& 0.076& 0.121& 0.154& 0.138& 0.121& 0.118& 0.120\\
		MvMDAp& 0.129& 0.173& 0.151& 0.159& 0.355& 0.257& 0.131& 0.220& 0.175\\
		MLDAp& 0.115& 0.108& 0.111& 0.234& 0.276& 0.255& 0.163& 0.164& 0.163\\
		GMAp& 0.130& 0.168& 0.149& 0.168& 0.337& 0.253& 0.134& 0.180& 0.157\\
		MvDA-CCAp& 0.136& 0.149& 0.143& 0.269& 0.400& 0.335& 0.151& 0.217& 0.184\\\hline
		
		DMCCA& 0.132& 0.118& 0.125& 0.115& 0.220& 0.168& 0.121& 0.156& 0.138\\
		DMvOPLS& 0.153& 0.161& 0.157& 0.406& \textbf{0.411}& 0.409& 0.198& 0.236& 0.217\\
		DMvDA& \textbf{0.155}& 0.162& \textbf{0.158}& 0.420& 0.399& 0.410& 0.210& 0.223& 0.217\\
		DMvDA-VC& \textbf{0.155}& 0.162& \textbf{0.158}& 0.420& 0.399& 0.410& 0.210& 0.223& 0.217\\
		DMvLDA& 0.133& 0.120& 0.126& 0.296& 0.290& 0.293& 0.129& 0.161& 0.145\\
		DMvMDA& 0.112& 0.173& 0.142& 0.376& 0.340& 0.358& 0.141& 0.226& 0.184\\
		DMLDA& 0.131& 0.117& 0.124& 0.229& 0.259& 0.244& 0.175& 0.179& 0.177\\
		DGMA& 0.122& 0.179& 0.151& 0.261& 0.276& 0.268& 0.146& 0.179& 0.162\\
		DMvDA-CCA& 0.153& \textbf{0.163}& \textbf{0.158}& \textbf{0.427}& 0.409& \textbf{0.418}& \textbf{0.213}& \textbf{0.239}& \textbf{0.226}\\\hline
	\end{tabular}
\end{table}

\subsubsection{Parameter sensitivity analysis}
In Section \ref{sec:reformulation}, we show that methods MvDA-VC, MLDA, GMA, MvDA-CCA and their nonlinear versions have another parameter $\lambda$. To investigate the impact of $\lambda$ on the four models, we repeat the experiments  in Section \ref{sec:overall-class} on  Caltech101-20 by fixing $k=50$ and varying $\lambda \in [ 10^{-4},10^{4}]$. Experimental results are shown in Fig. \ref{fig:params}. The four methods  show large variations on the original input data, but they behave less sensitive to $\lambda$ for reduced data using either  PCA or deep networks.

\subsection{Performance evaluation via text-image retrieval}

Text-image retrieval is used to evaluate MvOPLS and its variants on data sets whose two views are  texts and images, respectively. This task aims to retrieve image (text) from a database for a given  text (image) query. A correct retrieval is the one with the same class as the query. The mean average precision (mAP) score is the performance measurement for text-image retrieval, and it has been popularly used in \cite{pereira2012regularization,sharma2012generalized}. The parameters of all methods are the same as these used in Subsection \ref{sec:classification}. 

Table \ref{tab:retreival} shows the mAP scores of all compared methods on three data sets. We observe that 1) PCA is helpful on Pascal, but not for TVGraz and Wikipedia; 2) Deep variants outperform their counterparts. DMvDA-CCA shows the best performance in terms of mAP score over all three data sets. 3) DMvLDA does not show good performance for cross-modal retrieval as in classification (Subsection \ref{sec:classification}). These results demonstrate that MvOPLS with deep networks for learning nonlinear transformations is effective for certain models, but may not for all variants.

\section{Conclusion} \label{sec:conclusion}
In this paper, we propose a unified multi-view learning framework, which not only provides a deep understanding of many existing methods from the viewpoint of regularized least squares, but also motivates the development of new methods as well as their nonlinear counterparts with little additional effort. Extensive experiments in terms of two multi-view learning tasks validate the proposed framework, the two newly instantiated models, and the new deep variants.

The proposed framework provides appealing flexibility for designing effective models in a wide range of learning tasks. For example, the sparse CCA \cite{witten2009extensions,xu2019canonical} can be reformulated under the proposed framework with sparsity regularization over projection matrices. With the proposed framework, it becomes feasible to extend them for more than two views and nonlinear representations. Our framework can also be easily extended for other learning paradigm such as semi-supervised multi-view learning.

\end{document}